\documentclass{article} 
\usepackage{iclr2027_conference,times}
\usepackage{fancyhdr}
\usepackage{amsmath,amsfonts,bm}

\def\eqref#1{equation~\ref{#1}}

\def\1{\bm{1}}

\DeclareMathAlphabet{\mathsfit}{\encodingdefault}{\sfdefault}{m}{sl}
\SetMathAlphabet{\mathsfit}{bold}{\encodingdefault}{\sfdefault}{bx}{n}

\usepackage{titletoc}
\usepackage{tablefootnote}
\usepackage{hyperref}
\usepackage{url}
\usepackage{threeparttable}
\usepackage{comment}
\usepackage{booktabs}
\usepackage{longtable}
\usepackage[table]{xcolor}
\usepackage{multirow}
\usepackage{graphicx}
\usepackage{enumitem}
\usepackage{wrapfig}
\usepackage{wrapstuff}
\usepackage{tablefootnote}
\definecolor{geneiclblue}{HTML}{648FFF}
\colorlet{geneiclbluelight}{geneiclblue!20}
\definecolor{geneiclblue}{RGB}{232, 242, 252}

\title{GeneICL: A Tabular Foundation Model for Bulk Transcriptomics}

\author{
  Michael Bohl\thanks{Equal contribution.}~~\textsuperscript{1} \quad
  Alexander Theus\footnotemark[1]~~\textsuperscript{1,2} \quad
  David Wissel\textsuperscript{1} \quad
  Valentina Boeva\textsuperscript{1,3,4} \\
  \textsuperscript{1}ETH Zurich \quad
  \textsuperscript{2}Max Planck Institute for Intelligent Systems \\
  \textsuperscript{3}Swiss Institute of Bioinformatics \quad
  \textsuperscript{4}Université Paris Cité \\
  \texttt{\{mibohl,atheus,dwissel,vboeva\}@ethz.ch}
}

\iclrfinalcopy

\begin{document}
\maketitle
\fancyhf{}
\fancyhead[L]{}
\fancyfoot[C]{\thepage}
\renewcommand{\headrulewidth}{0.4pt}
\pagestyle{fancy}
\vspace{-0.5em}
\begin{abstract}
Gene expression is widely measured in biomedicine, yet clinical outcome prediction remains challenging due to high dimensionality, strong feature correlations, and limited labeled data. Large self-supervised transcriptomic foundation models often fail to outperform simple supervised baselines. Tabular foundation models offer an alternative through in-context learning, but are typically pretrained on generic synthetic data rather than transcriptomic structure. We ask whether transcriptomics-aware pretraining, rather than scale, is the missing ingredient. Towards this end, we introduce \textbf{GeneICL}, a 4.2M-parameter tabular foundation model combining a semi-synthetic pretraining prior built from measured bulk expression profiles with a parameter-efficient recurrent architecture. We further enable right-censored survival prediction via a training-free reduction to regression using Cox partial-likelihood residuals. We evaluate GeneICL on 80 clinical outcome-prediction tasks spanning classification, regression, and survival. Tabular foundation models consistently outperform self-supervised transcriptomic models, while GeneICL achieves the best overall rank among evaluated foundation models and tuned baselines. GeneICL does so with up to 387$\times$ fewer parameters, no gradient updates at inference, and predictions within seconds on a laptop CPU. Our code is available \href{https://github.com/BoevaLab/GeneICL}{here}\footnote{Github repository: \texttt{\href{https://github.com/BoevaLab/GeneICL}{https://github.com/BoevaLab/GeneICL}}.}.
\end{abstract}

\section{Introduction}
\vspace{-0.5em}
\begin{wrapfigure}[20]{R}{0.42\columnwidth}
    \centering
    \vspace{-1.2em}\includegraphics[width=\linewidth]
        {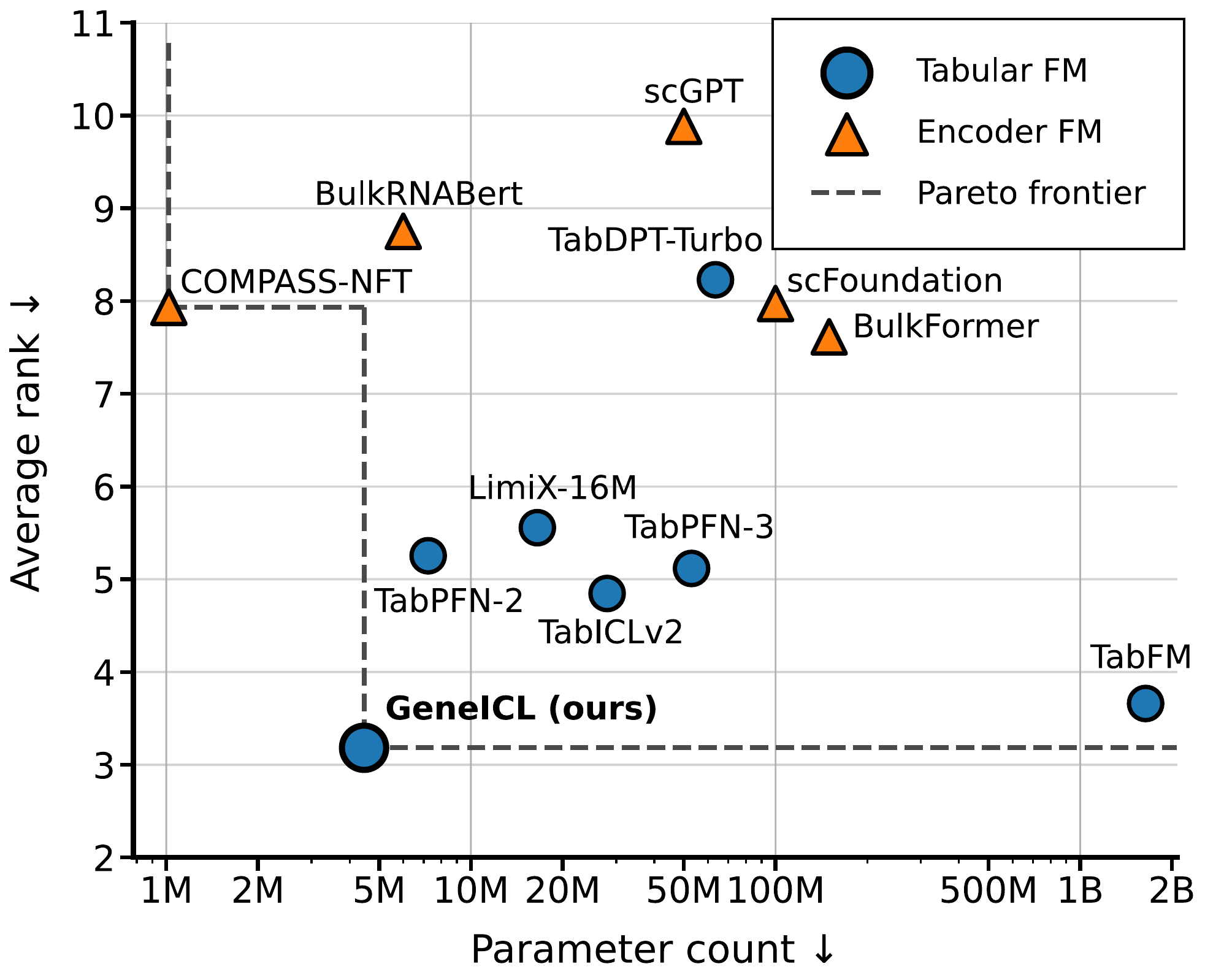}
    \caption{
    \textbf{Parameter–performance trade-off across foundation models.}
    Performance is the task-macro average rank across classification and regression tasks. The dashed line marks the Pareto frontier. Lower is better on both axes.
}
    \label{fig:parameter_efficiency}
\end{wrapfigure}

Gene expression provides a rich molecular readout of biological state and is routinely collected in clinical studies of disease progression and treatment response
~\citep{van2019rna}. Predictive models built from transcriptomic measurements, therefore, have the potential to support clinically relevant tasks such as forecasting treatment response, estimating prognosis, detecting transplant rejection, and enabling non-invasive diagnostics~\citep{byron2016translating}. At the same time, these prediction problems are statistically challenging. A typical transcriptomic dataset contains expression measurements for tens of thousands of genes but only tens to hundreds of labeled samples, placing many applications in the high-dimensional, low-sample-size regime~\citep{rezapour2026artificial}.

Recent transcriptomic foundation models use self-supervised pretraining on large collections of single-cell or bulk gene expression profiles~\citep{theodoris2023transfer,cui2024scgpt,kang2026bulkformer}. Inspired by the success of large-scale pretraining in other domains, these models aim to learn general-purpose representations that transfer to downstream tasks with limited labeled data. However, the extent to which self-supervised pretraining benefits supervised downstream prediction in transcriptomics remains debated. Independent evaluations have repeatedly found that simple supervised or statistical baselines can match or outperform substantially larger transcriptomic foundation models on downstream tasks~\citep{boiarsky2024deeper,ahlmann2025deep,kedzierska2025zero,hou2026unified}. This raises a basic question: if the goal is prediction from a small labeled gene expression cohort, \emph{is self-supervised representation learning the right pretraining objective?}

Tabular foundation models suggest a different approach. Models such as TabPFN, TabICL, TabDPT, LimiX, and TabFM are pretrained to solve supervised learning tasks through in-context learning~\citep{hollmann2025accurate,qu2026tabiclv2,ma2026tabdpt,wang2026limix,google2026tabfm}. At inference time, labeled training examples are provided as context, and predictions for new samples are produced without gradient-based fine-tuning. This paradigm is particularly attractive for transcriptomics, where each cohort naturally defines a small supervised learning problem. However, existing tabular foundation models are designed for generic tables rather than transcriptomic data, and therefore do not explicitly account for the statistical structure of gene expression. Transcriptomic datasets typically contain tens of thousands of strongly correlated features in the $n \ll p$ regime, with substantial redundancy arising from co-regulated gene expression programs. Recent methods such as TabPFN-Wide explicitly address the high-dimensional, low-sample-size setting, but remain designed for generic wide tables rather than transcriptomic data specifically~\citep{kolberg2025tabpfn}.

We introduce GeneICL, a parameter-efficient tabular foundation model purpose-built for transcriptomic prediction. GeneICL operates in principal component (PC) space and uses a recurrent, weight-tied in-context learner to predict query samples from labeled support examples. A single pretrained model supports both regression and classification. To adapt in-context pretraining to transcriptomics, GeneICL is pretrained on a semi-synthetic prior. Its expression profiles are sampled from a variational autoencoder trained on real bulk RNA-seq data, while its labels are generated by random synthetic functions of these profiles. The inputs thus carry the gene--gene correlation structure of real transcriptomes, while the labels remain as plentiful as in any synthetic prior.

We evaluate GeneICL on a benchmark of prognostic and diagnostic outcomes from bulk transcriptomic cohorts, spanning patient cohorts and cancer cell-line drug screens, including treatment response, prognosis, kidney transplant outcomes, and non-invasive diagnostics. We compare against tuned classical machine-learning methods, transcriptomic foundation models, and general-purpose tabular foundation models, and analyze the contribution of GeneICL's transcriptomics-specific prior and architecture.

Our contributions are threefold:
\begin{enumerate}[label=(\roman*)]
    \item \textbf{A translational outcome-prediction benchmark for bulk transcriptomics.} We systematically compare generic tabular and domain-specialized foundation models on prognostic and diagnostic tasks with direct translational relevance, moving beyond representation benchmarks such as cell-type or tissue-of-origin prediction.
    \item \textbf{GeneICL, a transcriptomics-specific tabular foundation model.} 
GeneICL pairs a semi-synthetic pretraining prior, whose covariates are generated from a model of over 500{,}000 experimental RNA-seq profiles, with a looped transformer that applies a single in-context block recurrently in place of a stack of distinct layers. With 4.2M parameters, GeneICL matches or outperforms general-purpose tabular foundation models with up to $387\times$ more parameters (Figure \ref{fig:parameter_efficiency}).
    \item \textbf{A training-free extension of tabular in-context learning to survival prediction.} We show that regressing on Cox partial-likelihood residuals lets a regression-capable tabular foundation model rank patients by risk without architectural changes or survival-specific pretraining. Our adaptation achieves the best overall rank against existing censoring adaptations for tabular foundation models, SurvPFN, a dedicated survival architecture, and tuned conventional baselines.
\end{enumerate}

\section{Related Work}

\subsection{Transcriptomic foundation models}
Large-scale pretraining has produced a growing family of transcriptomic foundation models. Single-cell models such as scGPT, scFoundation, Geneformer, and UCE train transformers with self-supervised objectives on large collections of cellular transcriptomes~\citep{cui2024scgpt,hao2024large,theodoris2023transfer,rosen2026universal}. This paradigm has since been extended to bulk RNA sequencing through models such as BulkRNABert and BulkFormer, as well as generative representation-learning approaches~\citep{gelard2024bulkrnabert,kang2026bulkformer,pande2026atlas}.

A common feature of these methods is that each transcriptome is encoded independently, without conditioning on other samples or their labels. Supervised downstream prediction therefore typically requires task-specific fine-tuning or an additional predictor. Moreover, independent evaluations have found that pretrained transcriptomic representations do not consistently outperform simpler supervised or statistical baselines~\citep{boiarsky2024deeper,ahlmann2025deep,kedzierska2025zero,hou2026unified}. GeneICL takes a different approach: labeled transcriptomes are processed jointly as an in-context learning problem, allowing predictions on a new dataset without task-specific parameter updates.

\subsection{Tabular foundation models}
Tabular foundation models are pretrained to solve supervised learning problems directly through in-context learning. TabPFN established this paradigm using millions of synthetic prediction tasks~\citep{muller2021transformers,hollmann2025accurate}. Subsequent work has extended it through pretraining on real tables, more diverse synthetic priors, and larger-scale architectures~\citep{ma2026tabdpt,zhang2026mitra,qu2026tabiclv2,grinsztajn2026tabpfn,google2026tabfm}. TabICL and TabICLv2, which GeneICL builds on, separate row encoding from in-context learning over rows and extend the framework to regression~\citep{qu2026tabiclv2}. Closest to our work, TabPFN-Wide adapts TabPFN-2 to high-dimensional data through continued pretraining on a high-dimensional prior~\citep{kolberg2025tabpfn}.

The connection between tabular foundation models and transcriptomics is beginning to emerge. General-purpose TabPFN and TabICL models have shown strong performance for perturbation prediction~\citep{palla2026tabular}, while BulkFormer representations combined with TabPFN have been applied to zero-shot prediction from bulk RNA-seq~\citep{tingle2026combining}. These approaches either use general-purpose tabular models directly or pair them with an independently pretrained transcriptomic encoder. GeneICL instead adapts the pretraining distribution and input representation of the in-context learner to bulk transcriptomic prediction, within a substantially more compact architecture.

General-purpose tabular foundation models natively support classification and regression, whereas many clinical datasets contain right-censored survival outcomes. Prior adaptations to survival analysis rely on specialized architectures and survival-specific priors~\citep{bohm2026survpfn,seletkov2026survival} or target imputation~\citep{vieyra2026staying,lyu2026censoring}. Instead of imputing censored event times, we regress Cox partial-likelihood residuals, allowing GeneICL to predict survival zero-shot without structural modifications or dedicated retraining.

\subsection{Synthetic priors}
Prior-data fitted tabular models depend strongly on the distribution of tasks encountered during pretraining. TabPFN generates synthetic tables from structural causal models~\citep{hollmann2025accurate}, while later methods such as TabICLv2 and Mitra broaden the pretraining distribution using more diverse synthetic generators~\citep{qu2026tabiclv2,zhang2026mitra}. Recent work has further highlighted the importance of the synthetic prior for downstream behavior~\citep{turkmen2026towards}. In contrast, TabDPT demonstrates that tabular in-context models can also be pretrained on collections of real datasets~\citep{ma2026tabdpt}.

\subsection{Looped Transformers}
In-context prediction can be viewed as learning an algorithm that maps labeled examples and a query to a prediction~\citep{akyurek2022learning,von2023transformers}. Since many learning algorithms are iterative, recurrent transformer architectures repeatedly apply shared layers across depth~\citep{dehghani2018universal}. \citet{yang2024looped} applied this idea specifically to in-context data fitting, arguing that recurrence provides an inductive bias aligned with iterative learning algorithms.

More recently, \citet{balef2026one} found substantial redundancy across the depth of tabular in-context learners and showed that repeatedly applying a single transformer layer can retain comparable predictive performance with only a fraction of the original parameters.

\section{GeneICL}
\subsection{Problem formulation}

We consider supervised prediction from bulk gene expression in the in-context learning setting. Each task consists of a labeled support set
\begin{equation}
\mathcal{S}
=
\left\{
\left(\mathbf{x}_i, y_i\right)
\right\}_{i=1}^{n_s},
\qquad
\mathbf{x}_i \in \mathbb{R}^{G},
\end{equation}
and an unlabeled query set
\begin{equation}
\mathcal{Q}
=
\left\{
\mathbf{x}_j
\right\}_{j=1}^{n_q},
\end{equation}
where $\mathbf{x}$ denotes a bulk gene expression profile over $G$ genes. The target $y$ is categorical for classification tasks and continuous for regression tasks.

Given the support examples, GeneICL predicts the labels of all query samples as
\begin{equation}\label{eq:icl_func}
\hat{\mathbf{y}}_{Q}
=
f_{\theta}
\left(
\mathbf{X}_{S},
\mathbf{y}_{S},
\mathbf{X}_{Q},
\tau
\right),
\end{equation}
where $\tau \in \{\mathrm{clf}, \mathrm{reg}\}$ denotes the task type and $\theta$ are the pretrained model parameters. The same pretrained model is used for both classification and regression through separate output heads.

Crucially, $\theta$ is kept fixed for every downstream task. GeneICL, therefore, adapts to a new dataset solely through the labeled examples provided in context and requires no task-specific parameter updates.

\begin{figure}[t]
    \includegraphics[width=0.95\linewidth]{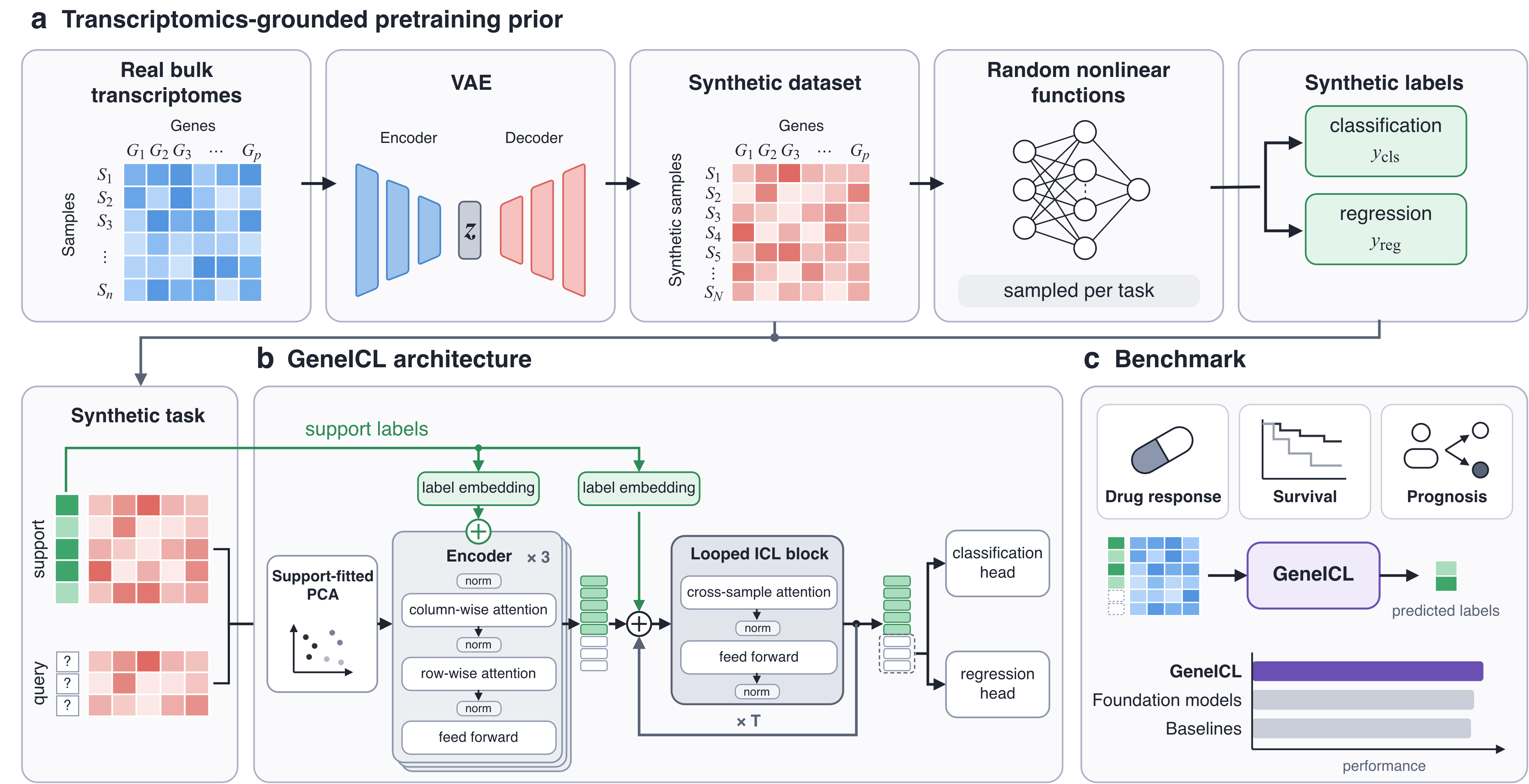}
    \caption{\textbf{Overview of GeneICL.} \textbf{(a)} Transcriptomics-grounded pretraining prior. A variational autoencoder (VAE) is trained on unlabeled real bulk transcriptomes. The decoder is used to generate synthetic expression datasets. A randomly sampled nonlinear function maps the individual profiles to a classification or regression label for every sample. Generated datasets are split into a labeled support set and a query set, forming one synthetic pretraining task. \textbf{(b)} GeneICL architecture. Transcriptomic profiles are projected with the support-fitted PCA and fed into an encoder. A label embedding of the support labels is added to the support tokens in the encoder and in each iteration of the weight-tied looped in-context-learning (ICL) transformer block. Classification and regression heads read out predictions from the query tokens. \textbf{(c)} Benchmark.  GeneICL is compared against tabular and transcriptomic foundation models and conventional baselines (performance shown schematically, see Table \ref{tab:main_results}).}\label{fig:overview}
\end{figure}
\subsection{Architecture}
\label{sec:geneicl_architecture}
GeneICL adopts the two-stage encoder--ICL architecture and most of the hyperparameter setup of TabICLv2~\citep{qu2026tabiclv2}, with a few architectural modifications aimed at simplifying the model and improving efficiency. Specifically, we (a) replace full feature mixing with a row-attention bottleneck over \texttt{[CLS]} tokens, (b) replace the stack of distinct ICL layers with a single weight-tied block applied recurrently over fixed evidence, (c) omit positional encodings and feature grouping (Appendix ~\ref{app:arch_symmetry}), and (d) use a unified model with a shared trunk for classification and regression. As shown in Section~\ref{sec:ablation}, these changes substantially reduce model size and computational cost without sacrificing predictive performance. Full architectural and training hyperparameters are provided in Appendix Table~\ref{tab:geneicl_hyperparameters}.
\paragraph{Model input.} Each dataset is standardized and projected onto its PCs, with both transformations fitted on the support set alone. We use PC scores rather than raw gene features for two reasons:
\begin{enumerate}
    \item \textbf{Rank in the $n \ll p$ regime.} Bulk transcriptomic cohorts contain far fewer samples than genes ($n_s \ll G \approx 20{,}000$). After centering, $\mathrm{rank}(\mathbf{X}_S) \le n_s - 1$, so all support-set variance lies in a subspace of at most $n_s - 1$ dimensions. Projecting onto the PCs is therefore lossless on the support set, while reducing the number of feature tokens from $G$ to at most $n_s - 1$.
    \item \textbf{Inductive bias.} Linear models fitted on PC scores are strong baselines in several independent benchmarks of transcriptomic representations~\citep{hou2026unified,elmarakeby2025empirical}. Rather than requiring the transformer to learn this dimensionality reduction from noisy, highly correlated gene features, we build it into the input.
\end{enumerate}

We keep the leading PCs that together explain 90\% of the support-set variance, discarding low-variance directions  without hurting performance (see Table \ref{tab:ablations_holdout}), and embed each PC score into $\mathbb{R}^{d}$ ($d = 128$) with a linear projection. Support rows are then augmented with label embeddings, which are learned class tokens for classification and linear projections of the normal-score-transformed target for regression.

\paragraph{Encoder.} As in TabFM~\citep{google2026tabfm}, the encoder alternates between column- and row-attention across three transformer blocks (see Figure \ref{fig:overview}b). Column attention mixes across samples within each feature using four inducing tokens that query only support rows. Row attention mixes information across features within each sample through four learned \texttt{[CLS]} tokens that act as an information bottleneck: features attend exclusively to and from these tokens, bypassing quadratic feature--feature attention without compromising performance (Section~\ref{sec:ablation}). After the final block, the four 128-dimensional \texttt{[CLS]} tokens per sample are normalized and concatenated into a 512-dimensional row representation. We denote the resulting support and query representations by
$Z_{\mathrm{supp}} \in \mathbb{R}^{n_s \times 512}$ and
$Z_{\mathrm{query}} \in \mathbb{R}^{n_q \times 512}$, respectively;
support representations additionally receive an additive second label embedding.

\paragraph{Looped in-context learner.} Inspired by work linking recurrence to in-context learning~\citep{yang2024looped,gatmiry2024can}, we replace the stack of distinct ICL layers in TabICLv2 with a single shared Transformer block applied recurrently. While prior tabular models have explored weight sharing across depth~\citep{balef2026one}, we follow input-injected recurrent architectures~\citep{geiping2026scaling} and separate fixed task evidence from an evolving hidden state, which yields stable extrapolation beyond the training depth (Appendix~\ref{app:arch_looping}).

Let $\mathbf{E}=[\mathbf{0};\mathbf{Z}_{\mathrm{supp}};\mathbf{Z}_{\mathrm{query}}]$ denote the fixed evidence and initialize
\begin{equation}
\mathbf{H}^{(0)}=[\mathbf{T}_{\mathrm{think}};\mathbf{0};\mathbf{0}],
\end{equation}
where $\mathbf{T}_{\mathrm{think}}$ contains eight learned thinking representations. We then iterate
\begin{equation}
\mathbf{H}^{(t+1)}=f_\theta\left(\mathbf{H}^{(t)}+\mathbf{E}\right),
\qquad t=0,\ldots,T-1.
\end{equation}
Keys and values are restricted to thinking and support positions, so queries attend to the evolving reasoning state and labeled support set but never to one another. We use $T=8$ steps; final query states are layer-normalized and passed to the corresponding classification or regression head.

\subsection{Transcriptomics-grounded pretraining prior}

Tabular foundation models are commonly pretrained either on fully synthetic tasks sampled from hand-designed generative priors~\citep{hollmann2025accurate,qu2026tabiclv2}, or directly on collections of real-world tables~\citep{ma2026tabdpt}. GeneICL takes an intermediate approach. We construct a semi-synthetic prior. Input features are sampled from a generative model of real expression profiles, and prediction targets are generated synthetically. This preserves the domain-specific covariate structure while allowing an effectively unlimited number of 
supervised tasks to be sampled during pretraining.

\paragraph{Transcriptomic data distribution.}
We derive the input distribution from ARCHS4~\citep{lachmann2018massive}, a large-scale collection of uniformly processed public RNA-seq experiments from GEO and SRA. We use \texttt{human\_gene\_v2.latest} (upstream build of 19 May 2026), comprising 547{,}829 human RNA-seq profiles after preprocessing and removal of all profiles contributing to benchmark tasks, represented over $G=20{,}021$ genes. The resulting corpus spans diverse tissues, cell types, diseases, and experimental conditions.

Rather than sampling these profiles directly during GeneICL pretraining, we first fit a variational autoencoder (VAE)~\citep{kingma2013auto} to their expression distribution. See Appendix~\ref{app:vae} for VAE details. Let $D_{\phi}$ denote the learned decoder. Synthetic expression profiles are generated as
\begin{equation}
    \mathbf{z}_i \sim \mathcal{N}(\mathbf{0},\mathbf{I}),
    \qquad
    \mathbf{x}_i = D_{\phi}(\mathbf{z}_i),
    \qquad i=1,\ldots,n.
\end{equation}
This allows us to sample an effectively unlimited number of expression datasets while retaining structure learned from real transcriptomic data, rather than repeatedly drawing from a finite set of observed profiles. The resulting matrix
$\mathbf{X}=[\mathbf{x}_1,\ldots,\mathbf{x}_n]^\top$
therefore defines a transcriptomics-grounded pretraining distribution rather than a generic synthetic tabular prior. As shown in Section~\ref{sec:ablation}, pretraining on VAE-generated profiles also yields slightly better mean performance than using real ARCHS4 profiles directly.

\paragraph{Synthetic task generation.}
For each generated expression dataset, we independently sample a supervised prediction problem. Targets are constructed in a principal-component representation of $\mathbf{X}$, matching the representation used by GeneICL downstream. Let $\mathbf{u}_i$ denote the whitened PCA representation of sample $i$. We sample a sparse subset $\mathcal{J}$ of principal components, with selection probabilities weighted by their explained variance, and generate latent task scores
\begin{equation}
    s_{ik}
    =
    g_{\psi_k}\!\left(\mathbf{u}_{i,\mathcal{J}}\right),
    \qquad k=1,\ldots,K,
\end{equation}
where the functions $g_{\psi_k}$ are independently sampled MLPs with randomized architectures, nonlinearities, and weight scales. Resampling both $\mathcal{J}$ and $\psi_k$ for every task yields a broad distribution of prediction functions while preferentially placing task-relevant signal along prominent axes of transcriptomic variation.

Classification and regression targets are constructed from the same latent score functions. For binary classification, the first score $s_{i1}$ is thresholded at a randomly sampled quantile, and for multiclass classification with $K$ classes, the label is the index of the largest score,
\begin{equation}
    y_i^{\mathrm{clf}}
    =
    \arg\max_{k \in \{1,\ldots,K\}} s_{ik}.
\end{equation}
For regression, the first latent score $s_{i1}$ is used as the target and corrupted with task-dependent noise to obtain a randomly sampled target signal-to-noise ratio.

Every sampled dataset provides both categorical and continuous views, $\mathbf{y}^{\mathrm{clf}}$ and $\mathbf{y}^{\mathrm{reg}}$, of the same underlying synthetic task. Consequently, a single generated expression matrix can jointly supervise GeneICL's classification and regression capabilities. Further details of the task-function prior and sampling distributions are provided in Appendix~\ref{app:pretraining_prior}.

\subsection{Training objective}\label{sec:objective}
Each synthetic task provides paired categorical and continuous query labels, $\mathbf{y}^{\mathrm{clf}}_{Q}$ and $\mathbf{y}^{\mathrm{reg}}_{Q}$. During pretraining, the task type $\tau$ selects which label type is embedded in the support set, but both output heads are supervised on the query samples of every task. We minimize
\begin{equation}\label{eq:loss}
\mathcal{L}
=
\mathrm{CE}\bigl(\hat{\mathbf{P}}_{Q}, \mathbf{y}^{\mathrm{clf}}_{Q}\bigr)
+
\mathcal{L}_{\mathrm{rank}}\bigl(\hat{\mathbf{P}}_{Q}, \mathbf{y}^{\mathrm{clf}}_{Q}\bigr)
+
\mathrm{MSE}\bigl(\hat{\mathbf{y}}^{\mathrm{reg}}_{Q}, \mathbf{y}^{\mathrm{reg}}_{Q}\bigr),
\end{equation}
where $\hat{\mathbf{P}}_{Q}$ are the class probabilities returned by the classification head, $\hat{\mathbf{y}}^{\mathrm{reg}}_{Q}$ the predictions of the regression head, and CE and MSE the cross-entropy and mean squared error averaged over query samples. $\mathbf{y}^{\mathrm{reg}}_{Q}$ is standardized with support-set statistics.

The ranking term $\mathcal{L}_{\mathrm{rank}} = -\mathrm{AUC}_{\mathrm{binormal}}$ replaces the macro-averaged one-vs-rest AUROC by its binormal approximation, computed in closed form from the first two moments of positive and negative query scores, making it a parametric analogue of the histogram loss~\citep{ustinova2016learning}. For each class $c$, we model the log-probabilities $\log \hat{p}_{c}$ of positive (label $c$) and negative (any other label) query samples as Gaussians with means $\mu_c^{+}, \mu_c^{-}$ and standard deviations $\sigma_c^{+}, \sigma_c^{-}$, estimated per task, which gives the closed form
\begin{equation}\label{eq:auc_binormal}
\mathrm{AUC}_{\mathrm{binormal}}
=
\frac{1}{|\mathcal{C}|}
\sum_{c \in \mathcal{C}}
\Phi\!\left(
\frac{\mu_c^{+}-\mu_c^{-}}
{\sqrt{(\sigma_c^{+})^{2}+(\sigma_c^{-})^{2}}}
\right),
\end{equation}
where $\Phi$ is the standard normal CDF and $\mathcal{C}$ the set of classes present in the query set.

\subsection{Cox-based tabular survival prediction}\label{sec:cox}

For right-censored outcomes $(t_i, \delta_i)$, with observed time $t_i$ and event indicator $\delta_i \in \{0,1\}$, we regress on null-model martingale residuals~\citep{therneau1990martingale},
\begin{equation}\label{eq:martingale}
    r_i = \delta_i - \hat{H}_0(t_i),
    \qquad
    \hat{H}_0(t) = \sum_{\tau_k \le t} \frac{d_k}{n_k},
\end{equation}
where $d_k$ events occur among the $n_k$ support samples at risk at event time $\tau_k$. The residual $r_i$ is the gradient of the Cox partial log-likelihood with respect to the log-risk $f_i$ at $\mathbf{f} = \mathbf{0}$, so predicting it is a functional gradient step from the null model. As $\mathbf{r}_S$ depends only on the support labels, GeneICL uses it as an ordinary regression target, $\mathbf{f}_Q = f_{\theta}(\mathbf{X}_S, \mathbf{r}_S, \mathbf{X}_Q, \mathrm{reg})$, and $\mathbf{f}_Q$ ranks query patients by risk. Survival curves follow from the Breslow estimator (Appendix~\ref{app:survival}).

\section{Benchmark}
\label{sec:benchmark}

\begin{figure}[h]
    \includegraphics[width=\linewidth]{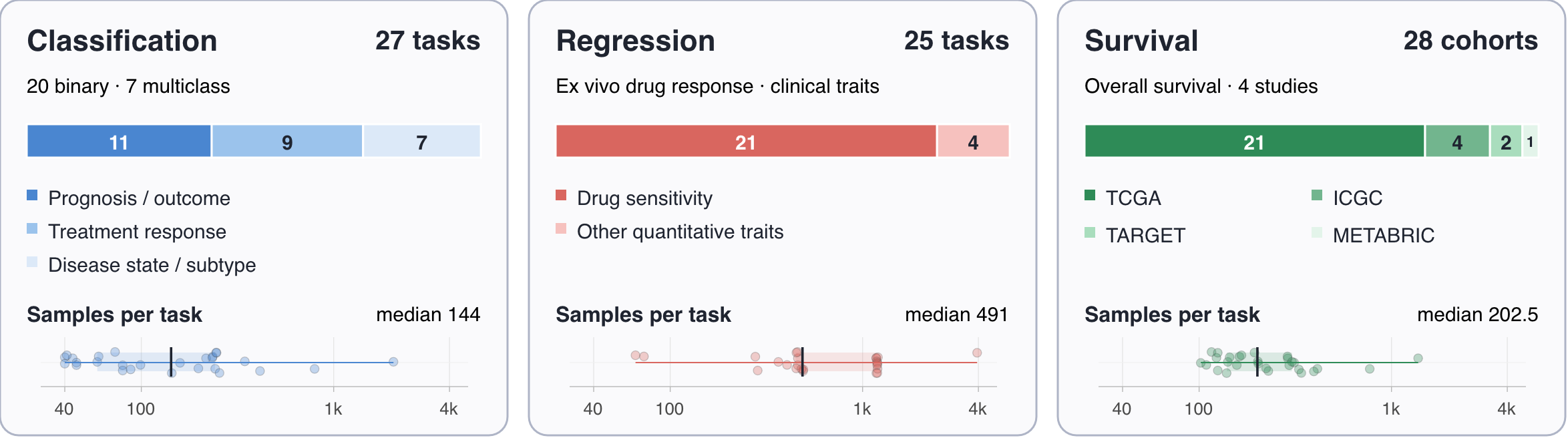}
    \centering
    \caption{\textbf{Benchmark overview.} The benchmark comprises 27 classification, 25 regression, and 28 survival tasks. Bars summarize endpoint categories or cohort sources. Sample-size distributions show individual tasks (dots), interquartile ranges (shading), and medians (ticks) on a shared logarithmic scale.}\label{fig:benchmark}
\end{figure}

\paragraph{Composition.}
Figure~\ref{fig:benchmark} summarizes the benchmark.
Classification comprises 20 binary and seven multiclass tasks,
covering treatment response, prognosis, and disease state or subtype.
Regression includes 21 drug-sensitivity endpoints and four other
quantitative traits. Multiple endpoints share underlying cohorts,
including ten drug-response tasks each from BeatAML and CREAMMIST~\citep{tyner2018functional, yingtaweesittikul_creammist_2023}.
Survival evaluation uses 28 cancer cohorts from TCGA, ICGC,
TARGET, and METABRIC through SurvBoard \citep{wissel2025survboard}.
Median sample sizes are 144, 491, and 202.5 for classification,
regression, and survival, respectively.
Dataset selection criteria and task-level details are provided
in Appendix~\ref{app:datasets}.

\paragraph{Evaluation protocol.}
Classification and regression methods use identical five-fold
cross-validation splits. Conventional baselines are tuned within
each training fold using 50 iterations of random search with
inner cross-validation; foundation models use fixed inference settings.
We evaluate classification using AUROC, AUPRC, and balanced accuracy,
and regression using Pearson's $r$, Spearman's $\rho$, and $R^2$.
Survival evaluation uses the first five predefined SurvBoard splits
and reports Antolini's C-index, integrated Brier score, and
D-calibration. Survival metrics are aggregated within each study
and then averaged across studies with equal study weighting.

\section{Results}
\subsection{Classification and Regression}

A clear separation emerges between model families (Table~\ref{tab:main_results}). Despite being pretrained directly on transcriptomic data, the self-supervised transcriptomic foundation models consistently trail the strongest general-purpose tabular foundation models and conventional baselines. In contrast, tabular foundation models are competitive with tuned supervised methods,
suggesting that pretraining directly for supervised prediction is well matched to the
high-dimensional, low-sample-size transcriptomic setting.

GeneICL further improves on this paradigm by combining the tabular in-context learning objective with a transcriptomics-grounded pretraining distribution. The standard GeneICL model achieves the second-best overall average rank across classification and regression, surpassed only by its inference-time ensemble. The ensemble achieves the best overall average rank, the highest AUROC and AUPRC, and the best result on all three regression metrics, while the tuned linear baseline remains strongest in balanced accuracy. Importantly, GeneICL contains only 4.24M parameters: approximately $387\times$ fewer than TabFM (1.65B), the strongest competing general-purpose tabular foundation model on regression. On a consumer laptop CPU, one cross-validation fold takes 7.5\,s on average, with peak memory below 5\,GB (Appendix~\ref{sec:cpu-runtime}).
\begin{table}[!t]
    \centering
    \caption{
        \textbf{Performance across 27 classification and 25 regression tasks.}
        Best and second-best results per metric are \textbf{bold} and \underline{underlined}, respectively.
        \emph{Avg.\ rank} is the mean rank across all six metrics and tasks (lower is better).
        $^{\dagger}$ denotes ensembled predictions.
        GeneICL$^{\dagger}$ averages predictions across PCA thresholds and support-set subsets
        (Appendix~\ref{app:arch_ensemble}).
        Parentheses for GeneICL denote the standard deviation over three seeds in units of the last decimal place.
    }
    \vspace{0.5em}
    \scriptsize
    \setlength{\tabcolsep}{5pt}
    \renewcommand{\arraystretch}{1.05}

    \begin{tabular}{lcccccccc}
        \toprule
        \textbf{Model}
        & \textbf{Params} $\downarrow$
        & \textbf{Avg.\ rank} $\downarrow$
        & \multicolumn{3}{c}{\textbf{Classification} $\uparrow$}
        & \multicolumn{3}{c}{\textbf{Regression} $\uparrow$} \\
        \cmidrule(lr){4-6}
        \cmidrule(lr){7-9}
        &
        &
        & \textbf{AUROC}
        & \textbf{AUPRC}
        & \textbf{Bal.\ Acc.}
        & \textbf{Pearson}
        & \textbf{Spearman}
        & $\mathbf{R^2}$ \\
        \midrule

        \multicolumn{9}{l}{
            \textit{Tabular foundation models}
        } \\
        \midrule

        \rowcolor{geneiclbluelight}
        GeneICL$^{\dagger}$
        & 4.24M
        & \textbf{4.64}
        & \textbf{0.742}(2)
        & \textbf{0.695}(4)
        & \underline{0.592}(6)
        & \textbf{0.528}(1)
        & \textbf{0.537}(1)
        & \textbf{0.274}(5) \\

        \rowcolor{geneiclbluelight}
        GeneICL
        & 4.24M
        & \underline{4.98}
        & \underline{0.741}(4)
        & 0.689(2)
        & 0.588(3)
        & \underline{0.521}(1)
        & 0.533(1)
        & \underline{0.271}(4) \\

        TabFM$^{\dagger}$
        & 1.65B
        & 5.65
        & 0.721
        & 0.686
        & 0.577
        & 0.515
        & \underline{0.535}
        & 0.268 \\

        TabICLv2$^{\dagger}$
        & 28.0M
        & 7.24
        & 0.712
        & 0.671
        & 0.584
        & 0.495
        & 0.519
        & 0.243 \\

        TabPFN-3$^{\dagger}$
        & 55.7M
        & 7.45
        & 0.726
        & 0.685
        & 0.575
        & 0.488
        & 0.508
        & 0.248 \\

        TabPFN-2$^{\dagger}$
        & 9M
        & 7.77
        & 0.720
        & 0.680
        & 0.566
        & 0.492
        & 0.504
        & 0.241 \\

        LimiX-16M$^{\dagger}$
        & 16.5M
        & 8.17
        & 0.705
        & 0.660
        & 0.541
        & 0.497
        & 0.514
        & 0.252 \\

        TabPFN-Wide\tablefootnote{TabPFN-Wide uses all genes and supports classification only. Regression results are therefore not reported.}
        & 7M
        & 8.25
        & 0.731
        & 0.692
        & 0.556
        & --
        & --
        & -- \\

        TabDPT-Turbo$^{\dagger}$
        & 63.5M
        & 12.43
        & 0.692
        & 0.650
        & 0.537
        & 0.394
        & 0.430
        & 0.100 \\

        \midrule
        \multicolumn{9}{l}{
            \textit{Encoder transcriptomics models}
        } \\
        \midrule

        BulkFormer
        & 147M
        & 11.86
        & 0.683
        & 0.636
        & 0.565
        & 0.437
        & 0.442
        & 0.188 \\

        COMPASS-NFT
        & 1.02M
        & 11.96
        & 0.684
        & 0.639
        & 0.564
        & 0.392
        & 0.387
        & 0.154 \\

        scFoundation
        & 100M
        & 12.35
        & 0.669
        & 0.630
        & 0.568
        & 0.424
        & 0.426
        & 0.164 \\

        BulkRNABert
        & 6.01M
        & 13.48
        & 0.670
        & 0.622
        & 0.542
        & 0.382
        & 0.390
        & 0.146 \\

        scGPT
        & 51.3M
        & 15.31
        & 0.613
        & 0.572
        & 0.516
        & 0.366
        & 0.363
        & 0.111 \\

        \midrule
        \multicolumn{9}{l}{
            \textit{Conventional baselines}
        } \\
        \midrule

        Linear (elastic-net)
        & --
        & 5.41
        & 0.735
        & \underline{0.694}
        & \textbf{0.621}
        & 0.500
        & 0.516
        & 0.262 \\

        CatBoost
        & --
        & 10.74
        & 0.705
        & 0.658
        & 0.547
        & 0.448
        & 0.460
        & 0.207 \\

        LightGBM
        & --
        & 12.20
        & 0.673
        & 0.633
        & 0.520
        & 0.439
        & 0.455
        & 0.202 \\

        MLP
        & --
        & 12.23
        & 0.684
        & 0.640
        & 0.566
        & 0.414
        & 0.431
        & 0.016 \\

        Random Forest
        & --
        & 12.73
        & 0.690
        & 0.638
        & 0.510
        & 0.432
        & 0.444
        & 0.186 \\

        \bottomrule
    \end{tabular}

    \label{tab:main_results}
\end{table}

\subsection{GeneICL can predict survival outcomes}\label{sec:survival_results}
\begin{figure}[t]
\centering
\includegraphics[width=0.55\linewidth, trim=0 10 0 10,
    clip]{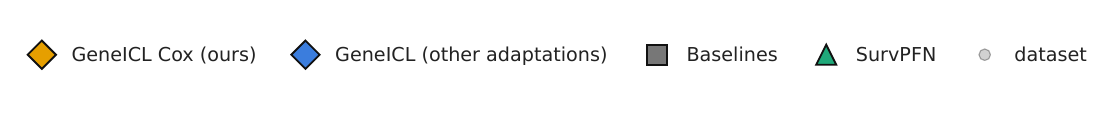}

\includegraphics[width=\linewidth]{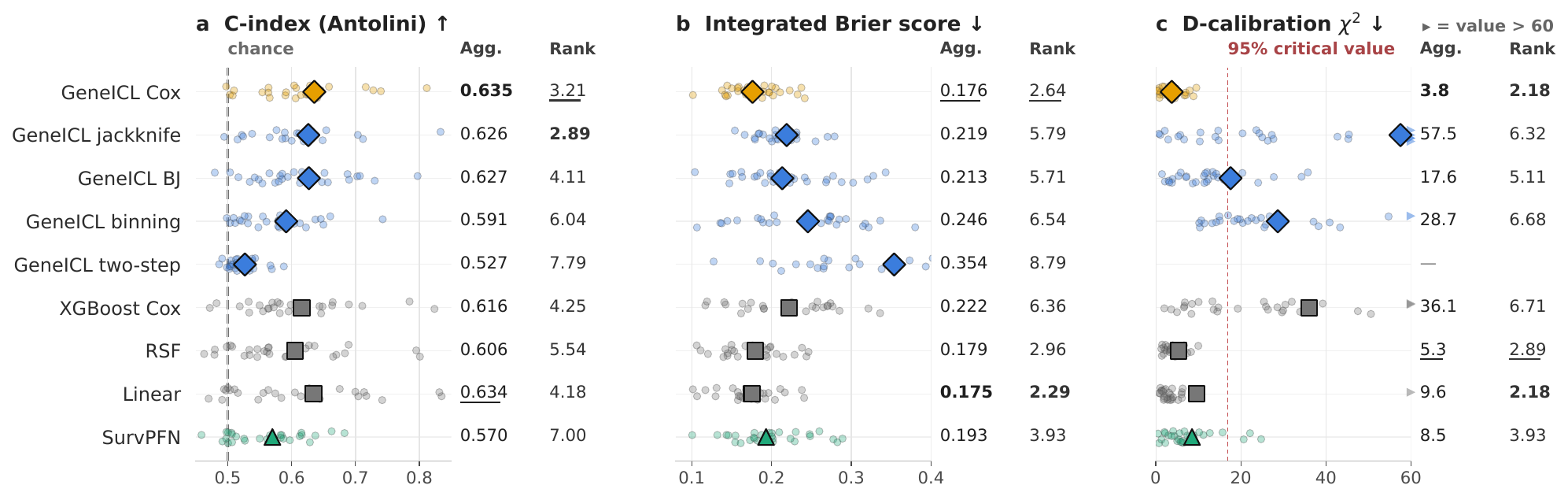}
\caption{
\textbf{Survival prediction across 28 SurvBoard datasets.}
We compare GeneICL with Cox residual fitting against alternative tabular survival adaptations, SurvPFN, and tuned baselines.
Panels show Antolini's C-index (\textbf{a}), integrated Brier score (\textbf{b}), and D-calibration $\chi^2$ (\textbf{c}).
Small points denote datasets; larger markers show study-weighted aggregates across TARGET, TCGA, METABRIC, and ICGC.
\emph{Agg.} reports this aggregate and \emph{Rank} the mean within-dataset rank; best and second-best values are \textbf{bold} and \underline{underlined}.
Dashed lines mark chance-level C-index ($0.5$) and the D-calibration threshold ($\chi^2=16.9$, $p=0.05$).
}
\label{fig:survival_results}
\end{figure}

\vspace{-4pt}
Clinical survival cohorts come with right-censored time-to-event outcomes, which GeneICL handles through the Cox partial-likelihood residual adaptation  (Section~\ref{sec:cox}). To isolate the effect of this adaptation, we first compare existing strategies for turning tabular foundation models into survival models with GeneICL as fixed backbone~\citep{vieyra2026staying,kim2026tabular,lyu2026censoring}. We also compare GeneICL-Cox against SurvPFN, a tabular foundation model with survival-specific architecture and pretraining~\citep{bohm2026survpfn}, and tuned survival baselines. In both settings, GeneICL-Cox achieves the highest aggregate C-index and best D-calibration, while remaining within $0.001$ of the best integrated Brier score (Figure~\ref{fig:survival_results}), without any survival-specific pretraining.

\subsection{Ablation study}\label{sec:ablation}

We first examine whether GeneICL's architectural simplifications come at a cost in predictive performance. Replacing the single recurrent ICL block with eight untied blocks increases the parameter count from 4.24M to 19.25M, yet does not improve any of the evaluated metrics (Table~\ref{tab:ablations}). Similarly, replacing the bottleneck feature attention with full self-attention in the encoder increases the feature-axis complexity from $\mathcal{O}(n)$ to $\mathcal{O}(n^2)$ with matched or lower performance. We further analyze the recurrent computation in Appendix~\ref{app:arch_looping}, where we vary the number of recurrence steps.

The pretraining distribution mainly affects regression. Replacing the VAE-generated covariates with either the generic TabICLv2 prior or the real ARCHS4 profiles underlying the VAE leaves classification ranking metrics largely unchanged but lowers all regression metrics ($R^2$ 0.238 and 0.248 vs.\ 0.271, see Table~\ref{tab:ablations}). Generating covariates with the VAE therefore costs no performance relative to real data, while providing an unlimited, leakage-free stream of novel expression profiles.

Finally, we investigate inference-time ensembling strategies in Appendix~\ref{app:arch_ensemble}. Across several alternative views of the same prediction problem, generic strategies based on PCA thresholds, support subsets, gene subsampling, and input transformations provide modest gains on several metrics, whereas transcriptomics-specific ensembles based on highly variable genes or Hallmark pathways do not consistently improve performance.

\begin{table}[t]
    \centering
    \caption{
        \textbf{Ablation of the main GeneICL design choices}.  \emph{Attn.} denotes the asymptotic cost of the feature-axis attention. Parentheses report the standard deviation over three seeds in units of the last decimal place. Best and second-best results are bold and underlined, respectively.
    }
    \vspace{0.5em}
    \scriptsize
    \setlength{\tabcolsep}{3pt}
    \renewcommand{\arraystretch}{1.05}

    \begin{tabular}{lccccccccc}
        \toprule
\textbf{Variant}
& \textbf{Params} $\downarrow$
& \textbf{Attn.}
& \textbf{Avg.\ rank} $\downarrow$
& \multicolumn{3}{c}{\textbf{Classification} $\uparrow$}
& \multicolumn{3}{c}{\textbf{Regression} $\uparrow$} \\
\cmidrule(lr){5-7}
\cmidrule(lr){8-10}
& & & &
\textbf{AUROC}
& \textbf{AUPRC}
& \textbf{Bal.\ Acc.}
& \textbf{Pearson}
& \textbf{Spearman}
& $\mathbf{R^2}$ \\
\midrule
        
        \rowcolor{geneiclbluelight}
        GeneICL & 4.24M & $\mathcal{O}(n)$ & \textbf{2.56}
        & \textbf{0.741}(4) & \underline{0.689}(2) & 0.588(3)
        & \textbf{0.521}(1) & \textbf{0.533}(1) & \underline{0.271}(4) \\
        \midrule
        
        Untied (8 blocks) & 19.25M & $\mathcal{O}(n)$ & \underline{2.70}
        & \underline{0.740}(2) & \underline{0.689}(3) & \underline{0.589}(1)
        & \underline{0.517}(1) & \underline{0.531}(2) & 0.263(6) \\
        
        Full attention & 4.05M & $\mathcal{O}(n^2)$ & 2.85
        & 0.739(0) & 0.688(2) & 0.570(6)
        & 0.515(3) & 0.530(1) & \textbf{0.273}(3) \\
        
        TabICLv2 prior pretraining & 4.24M & $\mathcal{O}(n)$ & 3.55
        & 0.737(1) & \textbf{0.690}(1) & 0.576(3)
        & 0.496(6) & 0.514(5) & 0.238(7) \\
        
        Real expression data pretraining & 4.24M & $\mathcal{O}(n)$ & 3.33
        & 0.735(3) & 0.682(3) & \textbf{0.593}(11)
        & 0.505(3) & 0.517(4) & 0.248(4) \\
        \bottomrule
    \end{tabular}

    \label{tab:ablations}
\end{table}
\section{Discussion and Conclusion}
Transcriptomic foundation model evaluations have largely centered on representation benchmarks such as cell-type or tissue-of-origin prediction, with clinical outcomes covered by only a few tasks. We therefore built a benchmark of 80 translational prognostic and diagnostic tasks, to ask whether self-supervised pretraining helps where it matters clinically. Our results suggest it does not. GeneICL, our 4.24M-parameter tabular foundation model, pretrained directly for supervised prediction, outperforms larger transcriptomic foundation models and matches general-purpose tabular foundation models with up to $387\times$ more parameters (Table~\ref{tab:main_results}). Through Cox partial-likelihood residual regression, the same frozen model also predicts survival without retraining. Across all three outcome types, the pretraining objective, rather than model scale, appears to be what limits current models.

Supervised pretraining benefits from tasks that resemble transcriptomic cohorts. Replacing our prior with the generic TabICLv2 prior, or VAE samples with real profiles, lowers regression performance (Table~\ref{tab:ablations}), so realistic covariates matter more than fidelity to individual samples. Architecture matters less than data. 
One looped block matches eight untied blocks with under a quarter of their parameters and the resulting model runs in seconds on a laptop CPU. 


\paragraph{Limitations.}
Our prior is derived from bulk RNA-seq, so single-cell RNA-seq and other omics modalities remain untested. Every task is also evaluated within a single cohort, which leaves robustness to batch and platform shifts between support and query samples open. Finally, whether the small margin over linear baselines reflects a limit of what tens to hundreds of expression profiles can reveal, or a limit of current priors and architectures, is the question we consider most important to answer next.

\section*{Acknowledgments and Disclosure of Funding}
We thank Johannes Kirschner for helpful feedback and comments. Alexander Theus is funded by the Max Planck ETH Center for Learning Systems.
\vspace{0.1em}
\section*{Code and data availability}

The ARCHS4 data we used to pretrain our VAE is available at \url{https://archs4.org/download} (version \textit{human\_gene\_v2.latest} from 5.7.2026). Most benchmark datasets are available at GEO (\url{https://www.ncbi.nlm.nih.gov/geo/}) under the accession numbers listed in Table \ref{tab:datasets}. BeatAML expression data and labels were downloaded from \url{https://github.com/biodev/beataml2.0_data}, GTEx v10 RNA-seq counts were downloaded from
\url{https://storage.googleapis.com/adult-gtex/bulk-gex/v10/rna-seq/GTEx_Analysis_v10_RNASeQCv2.4.2_gene_reads.gct.gz};
sample attributes and subject phenotypes were downloaded from
\url{https://storage.googleapis.com/adult-gtex/annotations/v10/metadata-files/GTEx_Analysis_v10_Annotations_SampleAttributesDS.txt}
and
\url{https://storage.googleapis.com/adult-gtex/annotations/v10/metadata-files/GTEx_Analysis_v10_Annotations_SubjectPhenotypesDS.txt},
respectively.
TARGET RNA-seq data were obtained from the NCI Genomic Data Commons at \url{https://portal.gdc.cancer.gov/projects/TARGET-NBL}. 
Cancer cell line profiles were downloaded from Model Passports \url{https://cellmodelpassports.sanger.ac.uk/downloads} (\textit{rnaseq\_merged\_20260323.zip}), with drug response labels from \url{https://creammist.mtms.dev/doc/bulk/}. The GeneICL source code and model weights for all three random seeds are available at \url{https://github.com/BoevaLab/GeneICL}.

\bibliography{iclr2027_conference}

@article{kang2026bulkformer,
  title={BulkFormer: A large-scale foundation model for bulk transcriptomes},
  author={Kang, Boming and Fan, Rui and Yi, Meizheng and Cui, Chunmei and Cui, Qinghua},
  journal={Cell Systems},
  year={2026},
  publisher={Elsevier}
}

@article{cui2024scgpt,
  title={scGPT: toward building a foundation model for single-cell multi-omics using generative AI},
  author={Cui, Haotian and Wang, Chloe and Maan, Hassaan and Pang, Kuan and Luo, Fengning and Duan, Nan and Wang, Bo},
  journal={Nature methods},
  volume={21},
  number={8},
  pages={1470--1480},
  year={2024},
  publisher={Nature Publishing Group US New York}
}

@article{hao2024large,
  title={Large-scale foundation model on single-cell transcriptomics},
  author={Hao, Minsheng and Gong, Jing and Zeng, Xin and Liu, Chiming and Guo, Yucheng and Cheng, Xingyi and Wang, Taifeng and Ma, Jianzhu and Zhang, Xuegong and Song, Le},
  journal={Nature methods},
  volume={21},
  number={8},
  pages={1481--1491},
  year={2024},
  publisher={Nature Publishing Group US New York}
}

@article{rosen2026universal,
  title={Universal cell embedding provides a foundation model for cell biology},
  author={Rosen, Yanay and Roohani, Yusuf and Agrawal, Ayush and Samotor{\v{c}}an, Leon and Tabula Sapiens Consortium and Quake, Stephen R and Leskovec, Jure},
  journal={Nature},
  pages={1--9},
  year={2026},
  publisher={Nature Publishing Group UK London}
}

@article{qu2026tabiclv2,
  title={TabICLv2: A better, faster, scalable, and open tabular foundation model},
  author={Qu, Jingang and Holzmüller, David and Varoquaux, Ga{\~A}l and Le Morvan, Marine},
  journal={arXiv preprint arXiv:2602.11139},
  year={2026}
}

@article{wang2026limix,
  title={LimiX-2M: Mitigating Low-Rank Collapse and Attention Bottlenecks in Tabular Foundation Models},
  author={Wang, Yuanrui and Zhang, Xingxuan and Yu, Han and Hao, Mingchao and Ren, Gang and Yuan, Hao and Mao, Li and Zhang, Yunjia and Yuan, Chun and Cui, Peng},
  journal={arXiv preprint arXiv:2606.04485},
  year={2026}
}

@article{grinsztajn2026tabpfn,
  title={Tabpfn-3: Technical report},
  author={Grinsztajn, L{\'e}o and Fl{\"o}ge, Klemens and Key, Oscar and Birkel, Felix and Jund, Philipp and Roof, Brendan and Manium, Mihir and Hoo, Shi Bin and B{\"u}hler, Magnus and Garg, Anurag and others},
  journal={arXiv preprint arXiv:2605.13986},
  year={2026}
}

@inproceedings{yang2024looped,
  title={Looped transformers are better at learning learning algorithms},
  author={Yang, Liu and Lee, Kangwook and Nowak, Robert and Papailiopoulos, Dimitris},
  booktitle={International conference on learning representations},
  volume={2024},
  pages={42195--42214},
  year={2024}
}

@article{balef2026one,
  title={Is One Layer Enough? Understanding Inference Dynamics in Tabular Foundation Models},
  author={Balef, Amir Rezaei and Koshil, Mykhailo and Eggensperger, Katharina},
  journal={arXiv preprint arXiv:2605.06510},
  year={2026}
}

@article{muller2021transformers,
  title={Transformers can do bayesian inference},
  author={M{\"u}ller, Samuel and Hollmann, Noah and Arango, Sebastian Pineda and Grabocka, Josif and Hutter, Frank},
  journal={arXiv preprint arXiv:2112.10510},
  year={2021}
}

@article{lachmann2018massive,
  title={Massive mining of publicly available RNA-seq data from human and mouse},
  author={Lachmann, Alexander and Torre, Denis and Keenan, Alexandra B and Jagodnik, Kathleen M and Lee, Hoyjin J and Wang, Lily and Silverstein, Moshe C and Ma'ayan, Avi},
  journal={Nature communications},
  volume={9},
  number={1},
  pages={1366},
  year={2018},
  publisher={Nature Publishing Group UK London}
}

@article{tyner2018functional,
  title={Functional genomic landscape of acute myeloid leukaemia},
  author={Tyner, Jeffrey W and Tognon, Cristina E and Bottomly, Daniel and Wilmot, Beth and Kurtz, Stephen E and Savage, Samantha L and Long, Nicola and Schultz, Anna Reister and Traer, Elie and Abel, Melissa and others},
  journal={Nature},
  volume={562},
  number={7728},
  pages={526--531},
  year={2018},
  publisher={Nature Publishing Group UK London}
}

@article{pande2026atlas,
  title={An atlas-scale generative model for unified representation learning of bulk RNA-seq data},
  author={Pande, Amit and Uyar, Bora and Akalin, Altuna},
  journal={bioRxiv},
  pages={2026--06},
  year={2026},
  publisher={Cold Spring Harbor Laboratory}
}

@article{boiarsky2024deeper,
  title={Deeper evaluation of a single-cell foundation model},
  author={Boiarsky, Rebecca and Singh, Nalini M and Buendia, Alejandro and Amini, Ava P and Getz, Gad and Sontag, David},
  journal={Nature Machine Intelligence},
  volume={6},
  number={12},
  pages={1443--1446},
  year={2024},
  publisher={Nature Publishing Group UK London}
}

@article{van2019rna,
  title={RNA sequencing data: hitchhiker's guide to expression analysis},
  author={Van den Berge, Koen and Hembach, Katharina M and Soneson, Charlotte and Tiberi, Simone and Clement, Lieven and Love, Michael I and Patro, Rob and Robinson, Mark D},
  journal={Annual Review of Biomedical Data Science},
  volume={2},
  number={1},
  pages={139--173},
  year={2019},
  publisher={Annual Reviews}
}

@article{rezapour2026artificial,
  title={Artificial Intelligence in Bulk RNA-Seq: Challenges and Potential Solutions},
  author={Rezapour, Mostafa and Trefry, Stephanie V and Opoku, Lorreta A and Narayanan, Aarthi},
  journal={Computational and Structural Biotechnology Journal},
  volume={35},
  number={1},
  pages={0039},
  year={2026},
  publisher={AAAS}
}

@article{ahlmann2025deep,
  title={Deep-learning-based gene perturbation effect prediction does not yet outperform simple linear baselines},
  author={Ahlmann-Eltze, Constantin and Huber, Wolfgang and Anders, Simon},
  journal={Nature Methods},
  volume={22},
  number={8},
  pages={1657--1661},
  year={2025},
  publisher={Nature Publishing Group US New York}
}

@article{theodoris2023transfer,
  title={Transfer learning enables predictions in network biology},
  author={Theodoris, Christina V and Xiao, Ling and Chopra, Anant and Chaffin, Mark D and Al Sayed, Zeina R and Hill, Matthew C and Mantineo, Helene and Brydon, Elizabeth M and Zeng, Zexian and Liu, X Shirley and others},
  journal={Nature},
  volume={618},
  number={7965},
  pages={616--624},
  year={2023},
  publisher={Nature Publishing Group UK London}
}

@article{kedzierska2025zero,
  title={Zero-shot evaluation reveals limitations of single-cell foundation models},
  author={Kedzierska, Kasia Z and Crawford, Lorin and Amini, Ava P and Lu, Alex X},
  journal={Genome Biology},
  volume={26},
  number={1},
  pages={101},
  year={2025},
  publisher={Springer}
}

@article{hou2026unified,
  title={A unified framework enables accessible deployment and comprehensive benchmarking of single-cell foundation models},
  author={Hou, Siyu and Yang, Penghui and Ma, Wenjing and Wang, Jade Xiaoqing and Zhou, Xiang},
  journal={bioRxiv},
  year={2026}
}

@article{kolberg2025tabpfn,
  title={Tabpfn-wide: Continued pre-training for extreme feature counts},
  author={Kolberg, Christopher and Kreuer, Jules and Huurdeman, Jonas and Ouaari, Sofiane and Eggensperger, Katharina and Pfeifer, Nico},
  journal={arXiv preprint arXiv:2510.06162},
  year={2025}
}

@article{hollmann2025accurate,
  title={Accurate predictions on small data with a tabular foundation model},
  author={Hollmann, Noah and M{\"u}ller, Samuel and Purucker, Lennart and Krishnakumar, Arjun and K{\"o}rfer, Max and Hoo, Shi Bin and Schirrmeister, Robin Tibor and Hutter, Frank},
  journal={Nature},
  volume={637},
  number={8045},
  pages={319--326},
  year={2025},
  publisher={Nature Publishing Group UK London}
}

@article{ma2026tabdpt,
  title={Tabdpt: Scaling tabular foundation models on real data},
  author={Ma, Junwei and Thomas, Valentin and Hosseinzadeh, Rasa and Labach, Alex and Cresswell, Jesse and Golestan, Keyvan and Yu, Guangwei and Caterini, Anthony L and Volkovs, Maks},
  journal={Advances in Neural Information Processing Systems},
  volume={38},
  pages={172692--172722},
  year={2025}
}

@article{tingle2026combining,
  title={Combining BulkFormer and TabPFN to predict post-transplant function from kidney biopsies during normothermic machine perfusion or cold storage},
  author={Tingle, Samuel J and Kourounis, Georgios and Kazerouni, Sofia and Spiers, Harry VM and Larraz, Miguel and Mehta, Maithili and MacMillan, Serena and Hosgood, Sarah A and Nicholson, Michael L and Sheerin, Neil S and others},
  journal={Scientific Reports},
  year={2026},
  publisher={Nature Publishing Group UK London}
}

@article{kim2026tabular,
  title={Tabular foundation models can do survival analysis},
  author={Kim, Da In and Lai, Wei Siang and Zhang, Kelly W},
  journal={arXiv preprint arXiv:2601.22259},
  year={2026}
}

@article{lyu2026censoring,
  title={A censoring-aware target interface for tabular foundation models in survival prediction},
  author={Lyu, Yue and Lin, Steven H and Huang, Xuelin and Li, Ziyi},
  journal={arXiv preprint arXiv:2607.09577},
  year={2026}
}

@article{vieyra2026staying,
  title={Staying Alive: Uncensored Survival Analysis with Tabular Foundation Models},
  author={Vieyra, Mariana Vargas},
  journal={arXiv preprint arXiv:2606.03689},
  year={2026}
}

@article{palla2026tabular,
  title={Tabular Foundation Models Are Competitive Cellular Perturbation Predictors Across Biological Scales},
  author={Palla, Giovanni and Hillsley, Alexander and Kim, Yang-Joon and Royer, Loic A},
  journal={bioRxiv},
  pages={2026--06},
  year={2026},
  publisher={Cold Spring Harbor Laboratory}
}

@article{turkmen2026towards,
  title={Towards Evaluating Data Priors for Tabular Foundation Models},
  author={T{\"u}rkmen, Zeynep and Kaya, K{\"u}r{\c{s}}at and Pfefferle, Alexander and Hutter, Frank},
  journal={arXiv preprint arXiv:2606.29241},
  year={2026}
}

@article{akyurek2022learning,
  title={What learning algorithm is in-context learning? investigations with linear models},
  author={Aky{\"u}rek, Ekin and Schuurmans, Dale and Andreas, Jacob and Ma, Tengyu and Zhou, Denny},
  journal={arXiv preprint arXiv:2211.15661},
  year={2022}
}

@inproceedings{von2023transformers,
  title={Transformers learn in-context by gradient descent},
  author={Von Oswald, Johannes and Niklasson, Eyvind and Randazzo, Ettore and Sacramento, Jo{\~a}o and Mordvintsev, Alexander and Zhmoginov, Andrey and Vladymyrov, Max},
  booktitle={International Conference on Machine Learning},
  pages={35151--35174},
  year={2023},
  organization={PMLR}
}

@article{dehghani2018universal,
  title={Universal transformers},
  author={Dehghani, Mostafa and Gouws, Stephan and Vinyals, Oriol and Uszkoreit, Jakob and Kaiser, {\L}ukasz},
  journal={arXiv preprint arXiv:1807.03819},
  year={2018}
}

@article{kingma2013auto,
  title={Auto-encoding variational bayes},
  author={Kingma, Diederik P and Welling, Max},
  journal={arXiv preprint arXiv:1312.6114},
  year={2013}
}

@misc{google2026tabfm,
  author={Kong, Weihao and Das, Abhimanyu},
  title={Introducing {TabFM}: A Zero-Shot Foundation Model for Tabular Data},
  howpublished={Google Research Blog},
  month=jun,
  year={2026},
  url={https://research.google/blog/introducing-tabfm-a-zero-shot-foundation-model-for-tabular-data/},
  note={Accessed: 2026-09-14}
}

@article{zhang2026mitra,
  title={Mitra: Mixed synthetic priors for enhancing tabular foundation models},
  author={Zhang, Xiyuan and Maddix Robinson, Danielle and Yin, Junming and Erickson, Nick and Ansari, Abdul Fatir and Han, Boran and Zhang, Shuai and Akoglu, Leman and Faloutsos, Christos and Mahoney, Michael and others},
  journal={Advances in neural information processing systems},
  volume={38},
  pages={15795--15840},
  year={2025}
}

@article{bohm2026survpfn,
  title={SurvPFN: Towards Foundation Models for Survival Predictions},
  author={B{\"o}hm, Samuel and Purucker, Lennart and Hutter, Frank and Schlosser, Pascal},
  journal={arXiv preprint arXiv:2606.04564},
  year={2026}
}

@article{seletkov2026survival,
  title={Survival In-Context: Amortized Bayesian Survival Analysis via Prior-Fitted Networks},
  author={Seletkov, Dmitrii and Hager, Paul and Kaissis, Georgios and Braren, Rickmer and Rueckert, Daniel and Rehms, Raphael},
  journal={arXiv preprint arXiv:2603.29475},
  year={2026}
}

@article{wissel2025survboard,
  title={SurvBoard: standardized benchmarking for multi-omics cancer survival models},
  author={Wissel, David and Janakarajan, Nikita and Grover, Aayush and Toniato, Enrico and Mart{\'\i}nez, Maria Rodr{\'\i}guez and Boeva, Valentina},
  journal={Briefings in Bioinformatics},
  volume={26},
  number={5},
  pages={bbaf521},
  year={2025},
  publisher={Oxford University Press}
}

@article{yang2022scbert,
  title={scBERT as a large-scale pretrained deep language model for cell type annotation of single-cell RNA-seq data},
  author={Yang, Fan and Wang, Wenchuan and Wang, Fang and Fang, Yuan and Tang, Duyu and Huang, Junzhou and Lu, Hui and Yao, Jianhua},
  journal={Nature machine intelligence},
  volume={4},
  number={10},
  pages={852--866},
  year={2022},
  publisher={Nature Publishing Group UK London}
}

@article{pearce2026transcriptformer,
  title={TranscriptFormer: A generative cell atlas across 1.5 billion years of evolution},
  author={Pearce, James D and Simmonds, Sara E and Mahmoudabadi, Gita and Krishnan, Lakshmi and Palla, Giovanni and Istrate, Ana-Maria and Tarashansky, Alexander and Nelson, Benjamin and Valenzuela, Omar and Li, Donghui and others},
  journal={Science},
  volume={393},
  number={6806},
  pages={aec8514},
  year={2026},
  publisher={American Association for the Advancement of Science}
}

@article{gatmiry2024can,
  title={Can looped transformers learn to implement multi-step gradient descent for in-context learning?},
  author={Gatmiry, Khashayar and Saunshi, Nikunj and Reddi, Sashank J and Jegelka, Stefanie and Kumar, Sanjiv},
  journal={arXiv preprint arXiv:2410.08292},
  year={2024}
}

@article{therneau1990martingale,
  title={Martingale-based residuals for survival models},
  author={Therneau, Terry M and Grambsch, Patricia M and Fleming, Thomas R},
  journal={Biometrika},
  volume={77},
  number={1},
  pages={147--160},
  year={1990},
  publisher={Oxford University Press}
}

@article{elmarakeby2025empirical,
  title={Empirical Evaluation of Single-Cell Foundation Models for Predicting Cancer Outcomes},
  author={Elmarakeby, Haitham and Roman, Ahmed and Johri, Shreya and Van Allen, Eliezer M},
  journal={bioRxiv},
  year={2025}
}

@article{ustinova2016learning,
  title={Learning deep embeddings with histogram loss},
  author={Ustinova, Evgeniya and Lempitsky, Victor},
  journal={Advances in neural information processing systems},
  volume={29},
  year={2016}
}

@article{gelard2024bulkrnabert,
  title={BulkRNABert: cancer prognosis from bulk RNA-seq based language models},
  author={G{\'e}lard, Maxence and Richard, Guillaume and Pierrot, Thomas and Courn{\`e}de, Paul-Henry},
  journal={bioRxiv},
  pages={2024--06},
  year={2024},
  publisher={Cold Spring Harbor Laboratory}
}

@article{van2019cell,
  title={Cell Model Passports—a hub for clinical, genetic and functional datasets of preclinical cancer models},
  author={van der Meer, Dieudonne and Barthorpe, Syd and Yang, Wanjuan and Lightfoot, Howard and Hall, Caitlin and Gilbert, James and Francies, Hayley E and Garnett, Mathew J},
  journal={Nucleic acids research},
  volume={47},
  number={D1},
  pages={D923--D929},
  year={2019},
  publisher={Oxford University Press}
}

@article{yingtaweesittikul2023creammist,
  title={CREAMMIST: an integrative probabilistic database for cancer drug response prediction},
  author={Yingtaweesittikul, Hatairat and Wu, Jiaxi and Mongia, Aanchal and Peres, Rafael and Ko, Karrie and Nagarajan, Niranjan and Suphavilai, Chayaporn},
  journal={Nucleic acids research},
  volume={51},
  number={D1},
  pages={D1242--D1248},
  year={2023},
  publisher={Oxford University Press}
}

@article{yang_genomics_2012,
    title = {Genomics of {Drug} {Sensitivity} in {Cancer} ({GDSC}): a resource for therapeutic biomarker discovery in cancer cells},
    volume = {41},
    copyright = {http://creativecommons.org/licenses/by-nc/3.0},
    issn = {0305-1048, 1362-4962},
    shorttitle = {Genomics of {Drug} {Sensitivity} in {Cancer} ({GDSC})},
    url = {http://academic.oup.com/nar/article/41/D1/D955/1059448/Genomics-of-Drug-Sensitivity-in-Cancer-GDSC-a},
    doi = {10.1093/nar/gks1111},
    language = {en},
    number = {D1},
    urldate = {2026-02-02},
    journal = {Nucleic Acids Research},
    author = {Yang, Wanjuan and Soares, Jorge and Greninger, Patricia and Edelman, Elena J. and Lightfoot, Howard and Forbes, Simon and Bindal, Nidhi and Beare, Dave and Smith, James A. and Thompson, I. Richard and Ramaswamy, Sridhar and Futreal, P. Andrew and Haber, Daniel A. and Stratton, Michael R. and Benes, Cyril and McDermott, Ultan and Garnett, Mathew J.},
    month = nov,
    year = {2012},
    pages = {D955--D961},
}

@article{barretina_cancer_2012,
    title = {The {Cancer} {Cell} {Line} {Encyclopedia} enables predictive modelling of anticancer drug sensitivity},
    volume = {483},
    issn = {0028-0836, 1476-4687},
    url = {https://www.nature.com/articles/nature11003},
    doi = {10.1038/nature11003},
    language = {en},
    number = {7391},
    urldate = {2026-02-03},
    journal = {Nature},
    author = {Barretina, Jordi and Caponigro, Giordano and Stransky, Nicolas and Venkatesan, Kavitha and Margolin, Adam A. and Kim, Sungjoon and Wilson, Christopher J. and Lehár, Joseph and Kryukov, Gregory V. and Sonkin, Dmitriy and Reddy, Anupama and Liu, Manway and Murray, Lauren and Berger, Michael F. and Monahan, John E. and Morais, Paula and Meltzer, Jodi and Korejwa, Adam and Jané-Valbuena, Judit and Mapa, Felipa A. and Thibault, Joseph and Bric-Furlong, Eva and Raman, Pichai and Shipway, Aaron and Engels, Ingo H. and Cheng, Jill and Yu, Guoying K. and Yu, Jianjun and Aspesi, Peter and De Silva, Melanie and Jagtap, Kalpana and Jones, Michael D. and Wang, Li and Hatton, Charles and Palescandolo, Emanuele and Gupta, Supriya and Mahan, Scott and Sougnez, Carrie and Onofrio, Robert C. and Liefeld, Ted and MacConaill, Laura and Winckler, Wendy and Reich, Michael and Li, Nanxin and Mesirov, Jill P. and Gabriel, Stacey B. and Getz, Gad and Ardlie, Kristin and Chan, Vivien and Myer, Vic E. and Weber, Barbara L. and Porter, Jeff and Warmuth, Markus and Finan, Peter and Harris, Jennifer L. and Meyerson, Matthew and Golub, Todd R. and Morrissey, Michael P. and Sellers, William R. and Schlegel, Robert and Garraway, Levi A.},
    month = mar,
    year = {2012},
    pages = {603--607},
}

@article{rees2016correlating,
  title={Correlating chemical sensitivity and basal gene expression reveals mechanism of action},
  author={Rees, Matthew G and Seashore-Ludlow, Brinton and Cheah, Jaime H and Adams, Drew J and Price, Edmund V and Gill, Shubhroz and Javaid, Sarah and Coletti, Matthew E and Jones, Victor L and Bodycombe, Nicole E and others},
  journal={Nature chemical biology},
  volume={12},
  number={2},
  pages={109--116},
  year={2016},
  publisher={Nature Publishing Group US New York}
}

@article{kugathasan2017prediction,
  title={Prediction of complicated disease course for children newly diagnosed with Crohn's disease: a multicentre inception cohort study},
  author={Kugathasan, Subra and Denson, Lee A and Walters, Thomas D and Kim, Mi-Ok and Marigorta, Urko M and Schirmer, Melanie and Mondal, Kajari and Liu, Chunyan and Griffiths, Anne and Noe, Joshua D and others},
  journal={The Lancet},
  volume={389},
  number={10080},
  pages={1710--1718},
  year={2017},
  publisher={Elsevier}
}

@article{hyams2019clinical,
  title={Clinical and biological predictors of response to standardised paediatric colitis therapy (PROTECT): a multicentre inception cohort study},
  author={Hyams, Jeffrey S and Thomas, Sonia Davis and Gotman, Nathan and Haberman, Yael and Karns, Rebekah and Schirmer, Melanie and Mo, Angela and Mack, David R and Boyle, Brendan and Griffiths, Anne M and others},
  journal={The Lancet},
  volume={393},
  number={10182},
  pages={1708--1720},
  year={2019},
  publisher={Elsevier}
}

@article{cancer2013genomic,
  title={Genomic and epigenomic landscapes of adult de novo acute myeloid leukemia},
  author={Cancer Genome Atlas Research Network},
  journal={New England Journal of Medicine},
  volume={368},
  number={22},
  pages={2059--2074},
  year={2013},
  publisher={Mass Medical Soc}
}

@article{bolouri2018molecular,
  title={The molecular landscape of pediatric acute myeloid leukemia reveals recurrent structural alterations and age-specific mutational interactions},
  author={Bolouri, Hamid and Farrar, Jason E and Triche Jr, Timothy and Ries, Rhonda E and Lim, Emilia L and Alonzo, Todd A and Ma, Yussanne and Moore, Richard and Mungall, Andrew J and Marra, Marco A and others},
  journal={Nature medicine},
  volume={24},
  number={1},
  pages={103--112},
  year={2018},
  publisher={Nature Publishing Group US New York}
}

@article{geiping2026scaling,
  title={Scaling up test-time compute with latent reasoning: A recurrent depth approach},
  author={Geiping, Jonas and McLeish, Sean and Jain, Neel and Kirchenbauer, John and Singh, Siddharth and Bartoldson, Brian and Kailkhura, Bhavya and Bhatele, Abhinav and Goldstein, Tom},
  journal={Advances in Neural Information Processing Systems},
  volume={38},
  pages={41340--41391},
  year={2025}
}

@inproceedings{qi2023survivaleval,
  title={SurvivalEVAL: A comprehensive open-source python package for evaluating individual survival distributions},
  author={Qi, Shi-ang and Sun, Weijie and Greiner, Russell},
  booktitle={Proceedings of the AAAI Symposium Series},
  volume={2},
  pages={453--457},
  year={2023}
}

@article{graf1999assessment,
  title={Assessment and comparison of prognostic classification schemes for survival data},
  author={Graf, Erika and Schmoor, Claudia and Sauerbrei, Willi and Schumacher, Martin},
  journal={Statistics in medicine},
  volume={18},
  number={17-18},
  pages={2529--2545},
  year={1999},
  publisher={Wiley Online Library}
}

@article{haider2020effective,
  title={Effective ways to build and evaluate individual survival distributions},
  author={Haider, Humza and Hoehn, Bret and Davis, Sarah and Greiner, Russell},
  journal={Journal of Machine Learning Research},
  volume={21},
  number={85},
  pages={1--63},
  year={2020}
}

@article{antolini2005time,
  title={A time-dependent discrimination index for survival data},
  author={Antolini, Laura and Boracchi, Patrizia and Biganzoli, Elia},
  journal={Statistics in medicine},
  volume={24},
  number={24},
  pages={3927--3944},
  year={2005},
  publisher={Wiley Online Library}
}

@article{polsterl2020scikit,
  title={scikit-survival: A Library for Time-to-Event Analysis Built on Top of scikit-learn},
  author={P{\"o}lsterl, Sebastian},
  journal={Journal of Machine Learning Research},
  volume={21},
  number={212},
  pages={1--6},
  year={2020}
}

@article{glmnet,
    title = {Regularization Paths for Generalized Linear Models via
      Coordinate Descent},
    author = {Jerome Friedman and Trevor Hastie and Robert Tibshirani},
    journal = {Journal of Statistical Software},
    year = {2010},
    volume = {33},
    number = {1},
    pages = {1--22},
    doi = {10.18637/jss.v033.i01},
  }

@article{glmnet_cox,
    title = {Regularization Paths for Cox's Proportional Hazards Model
      via Coordinate Descent},
    author = {Noah Simon and Jerome Friedman and Trevor Hastie and
      Robert Tibshirani},
    journal = {Journal of Statistical Software},
    year = {2011},
    volume = {39},
    number = {5},
    pages = {1--13},
    doi = {10.18637/jss.v039.i05},
  }

@article{rsfs,
 ISSN = {19326157},
 URL = {http://www.jstor.org/stable/30245111},
 author = {Hemant Ishwaran and Udaya B. Kogalur and Eugene H. Blackstone and Michael S. Lauer},
 journal = {The Annals of Applied Statistics},
 number = {3},
 pages = {841--860},
 publisher = {Institute of Mathematical Statistics},
 title = {Random Survival Forests},
 urldate = {2026-09-24},
 volume = {2},
 year = {2008}
}

@inproceedings{xgb,
author = {Chen, Tianqi and Guestrin, Carlos},
title = {XGBoost: A Scalable Tree Boosting System},
year = {2016},
isbn = {9781450342322},
publisher = {Association for Computing Machinery},
address = {New York, NY, USA},
url = {https://doi.org/10.1145/2939672.2939785},
doi = {10.1145/2939672.2939785},
booktitle = {Proceedings of the 22nd ACM SIGKDD International Conference on Knowledge Discovery and Data Mining},
pages = {785–794},
numpages = {10},
location = {San Francisco, California, USA},
series = {KDD '16}
}

@article{shen2026generalizable,
  title={Generalizable AI predicts immunotherapy outcomes across cancers and treatments},
  author={Shen, Wanxiang and Moon, Intae and Nguyen, Thinh H and Li, Michelle M and Huang, Yepeng and Nair, Nitya and Marbach, Daniel and Zitnik, Marinka},
  journal={Nature Medicine},
  pages={1--13},
  year={2026},
  publisher={Nature Publishing Group US New York}
}

@article{byron2016translating,
  title={Translating RNA sequencing into clinical diagnostics: opportunities and challenges},
  author={Byron, Sara A and Keuren-Jensen, Van and Kendall, R and Engelthaler, David M and Carpten, John D and Craig, David W},
  journal={Nature Reviews Genetics},
  volume={17},
  number={5},
  pages={257--271},
  year={2016},
  publisher={Nature Publishing Group}
}

@article{liberzon2015molecular,
  title={The molecular signatures database hallmark gene set collection},
  author={Liberzon, Arthur and Birger, Chet and Thorvaldsd{\'o}ttir, Helga and Ghandi, Mahmoud and Mesirov, Jill P and Tamayo, Pablo},
  journal={Cell systems},
  volume={1},
  number={6},
  pages={417--425},
  year={2015},
  publisher={Elsevier}
}

@article{bolstad2003comparison,
  title={A comparison of normalization methods for high density oligonucleotide array data based on variance and bias},
  author={Bolstad, Benjamin M and Irizarry, Rafael A and {\AA}strand, Magnus and Speed, Terence P.},
  journal={Bioinformatics},
  volume={19},
  number={2},
  pages={185--193},
  year={2003},
  publisher={Oxford University Press}
}

@article{brennecke2013accounting,
  title={Accounting for technical noise in single-cell RNA-seq experiments},
  author={Brennecke, Philip and Anders, Simon and Kim, Jong Kyoung and Ko{\l}odziejczyk, Aleksandra A and Zhang, Xiuwei and Proserpio, Valentina and Baying, Bianka and Benes, Vladimir and Teichmann, Sarah A and Marioni, John C and others},
  journal={Nature methods},
  volume={10},
  number={11},
  pages={1093--1095},
  year={2013},
  publisher={Nature Publishing Group US New York}
}

@article{schulz2026measurement,
  title={Measurement noise limits the advantage of nonlinear models over linear models in biomedical prediction},
  author={Schulz, Marc-Andre and Ritter, Kerstin},
  journal={arXiv preprint arXiv:2606.18420},
  year={2026}
}

@article{yingtaweesittikul_creammist_2023,
    title = {{CREAMMIST}: an integrative probabilistic database for cancer drug response prediction},
    volume = {51},
    issn = {1362-4962},
    shorttitle = {{CREAMMIST}},
    doi = {10.1093/nar/gkac911},
    language = {eng},
    number = {D1},
    journal = {Nucleic Acids Research},
    author = {Yingtaweesittikul, Hatairat and Wu, Jiaxi and Mongia, Aanchal and Peres, Rafael and Ko, Karrie and Nagarajan, Niranjan and Suphavilai, Chayaporn},
    month = jan,
    year = {2023},
    pmid = {36259664},
    pmcid = {PMC9825458},
    pages = {D1242--D1248},
}
\bibliographystyle{iclr2027_conference}

\newpage
\appendix
\startcontents[appendix]

\section*{Appendix Contents}
\printcontents[appendix]{}{1}{}
\newpage
\section{Evaluation datasets}
\label{app:datasets}

\paragraph{Scope.}
Transcriptomic foundation models are commonly validated on tasks that probe the quality of integrated embeddings, such as cell-type annotation~\citep{yang2022scbert,rosen2026universal} or tissue-of-origin prediction~\citep{pearce2026transcriptformer, pande2026atlas}. These tasks are informative about representation quality but have limited bearing on clinical decisions. Our benchmark instead consists of prognostic and diagnostic prediction tasks on bulk transcriptomic cohorts. It comprises 27 classification and 25 regression tasks (52 in total) and, separately, 28 survival tasks. Table~\ref{tab:datasets} lists every task with its source, target, sample size, and task type.

\paragraph{Task categories.}
The classification and regression tasks cover (i) response to drugs, vaccines, or other therapies, including ex-vivo drug sensitivity of patient-derived cells and cell lines, (ii) prognosis, such as recurrence or relapse, and (iii) non-invasive diagnosis from blood transcriptomes, together with a few tumor and tissue phenotypes (e.g. Ki67 index or telomere length). Survival prediction forms a fourth group and is evaluated under a separate protocol described below.

\paragraph{Data sources.}
Classification and regression tasks are drawn from three kinds of sources. The first are large clinical cohorts: TARGET~\citep{bolouri2018molecular}, TCGA~\citep{cancer2013genomic}, BeatAML~\citep{tyner2018functional}, and the pediatric inflammatory bowel disease cohorts PROTECT (ulcerative colitis) and RISK (Crohn's disease) ~\citep{hyams2019clinical, kugathasan2017prediction}. The second are preclinical pharmacogenomic screens including cell-line drug-response panels from GDSC, CCLE, and CTRP~\citep{yang_genomics_2012, barretina_cancer_2012, rees2016correlating} obtained in harmonized form from Cell Model Passports~\citep{van2019cell} and CREAMMIST~\citep{yingtaweesittikul2023creammist}. The third are smaller clinical cohorts deposited in the Gene Expression Omnibus (GEO), identified by mining GEO sample annotations for clinical outcome keywords (e.g., \emph{clinical}, \emph{therapy}, \emph{response}, \emph{diagnosis}, etc.). We only retained study designs in which expression was profiled at baseline, before the outcome was measured, and excluded cross-sectional studies. Most candidate studies were excluded because of their size (\(n\leq30\) patients, \(n < 5\) samples per class), because the sample was taken at the same time as the outcome was recorded rather than at baseline before it, or because their target was unpredictable (AUROC \(\leq 0.55\) or Spearman \(\rho \leq 0.1\) with TabPFN-3). We retained only longitudinal designs in which expression precedes the outcome, so that the prediction task is prospective.

\paragraph{Survival tasks.}
For survival prediction, we use SurvBoard~\citep{wissel2025survboard}, a benchmark that standardizes preprocessing, data splits, and evaluation for cancer survival models. It comprises 28 cancer datasets from four programs: TCGA, ICGC, TARGET, and METABRIC. SurvBoard contains clinical and multi-omic data, but we exclusively use the gene-expression modality.

{\setlength{\LTcapwidth}{\linewidth}
\begin{longtable}{l c l p{7.7cm}}
\caption{\textbf{Datasets used in the benchmark.} $n$ is the number of usable (non-NaN) rows (samples) for the
associated task. For datasets with several tasks of the same type (i.e. BeatAML, CMP), the range over tasks is given. Type: cls = classification, reg = regression.}
\label{tab:datasets} \\
\toprule
\textbf{Dataset} & \textbf{$n$} & \textbf{Type} & \textbf{Description} \\
\midrule
\endfirsthead
\multicolumn{4}{c}{\tablename\ \thetable{} -- continued} \\
\toprule
\textbf{Dataset} & \textbf{$n$} & \textbf{Type} & \textbf{Description} \\
\midrule
\endhead
\midrule
\multicolumn{4}{r}{\textit{continued on next page}} \\
\endfoot
\bottomrule
\endlastfoot
GSE107422 & 80 & cls & Systemic recurrence from colorectal cancer resection tissue\\
GSE116324 & 44 & cls & Bortezomib-induction response (response sustained $\geq$1\,yr) in multiple myeloma patients \\
GSE120622 & 80 & cls & Non-Small cell lung cancer (surgical tumor) relapse \\
GSE154261 & 73 & cls & 24-month recurrence of early stage non-muscle-invasive bladder cancer\\
GSE156699 & 88 & cls & platinum chemotherapy response (PFS $\geq$6\,mo) of high-grade serous ovarian cancer patients\\
GSE157657 & 59 & cls & Anti-tuberculosis therapy response (4-class) from whole blood\\
GSE176178 & 40 & cls & Durable vs.\ non-durable Bacillus Calmette-Guerin (immunotherapy) response in early stage bladder cancer patients \\
GSE198520 & 46 & cls & EULAR response to anti-TNF therapy (3-class) from rheumatoid arthritis synovium \\
GSE213346 & 40 & cls & Recalcitrant vs.\ stable disease status at 1 year from oral lichen planus biopsy \\
GSE278476 & 46 & cls & Immune responder status from peripheral blood mononuclear cells in a MUC1-vaccine prevention trial \\
GSE294705 & 60 & cls & High-frequency recurrence index in non-muscle-invasive bladder cancer patients post Bacillus Calmette-Guerin therapy \\
GSE316750 & 41 & cls & Biochemical recurrence from radical-prostatectomy tumor \\
GSE54460 & 99 & cls & Biochemical recurrence from radical-prostatectomy tissue \\
GSE109142 & 198 & cls & Week-4 clinical remission in pediatric ulcerative colitis \\
GSE112927 & 235 & cls & Death-censored kidney allograft loss from pre-transplant blood \\
GSE163882 & 217 & cls & pathological complete response to neoadjuvant chemotherapy from breast cancer pre-treatment biopsy \\
GSE185263 & 345 & cls & In-hospital mortality from early-sepsis whole blood \\
GSE243375 & 242 & cls & Pathological complete response to HER2+ breast neoadjuvant T-DM1+pertuzumab \\
GSE93624 & 245 & cls & 3-year complicated progression from pediatric Crohn's disease ileal biopsy \\
GSE175718 & 234 & cls & Kidney-transplant rejection type (4 classes) from blood \\
GSE107995 & 414 & cls & Future tuberculosis status (4 classes) from whole blood \\
GSE156902 & 796 & cls & GBM/brain-metastasis/MS/control classification (4-class) from tumor-educated platelets (peripheral blood) \\
GSE234297 & 144 & cls & ALS vs.\ healthy diagnosis from whole blood \\
GSE279480 & 255 & cls & CMV serostatus (positive/negative) from whole blood \\
GSE68086 & 246 & cls & Pan-cancer type vs.\ healthy classification (7-class) from tumor-educated platelets (peripheral blood)\\
TARGET & 159 & cls & INSS stage from pediatric neuroblastoma tumors \\
TARGET & 2056 & cls & relapse from pediatric tumors \\
GSE115525 & 73 & reg & Prednisolone LC50 drug sensitivity of B-lineage ALL primary cells \\
GSE157103 & 66 & reg & Hospital-free days (45-day window) from leukocytes of COVID-19 intensive care unit patients\\
BeatAML & 277--492 & reg & AML patient-derived ex-vivo drug sensitivity (AUC) for 10 drugs \\
GTEx & 3940 & reg & Telomere length from multi-tissue samples \\
GSE81538 & 405 & reg & Ki67 proliferation index from breast cancer \\
CMP\(^1\) & 1177--1202 & reg & IC50 drug sensitivity of pan-cancer cell lines for 10 drugs \\
TARGET & 453 & reg & Minimal residual disease level from pediatric tumors \\
\end{longtable}
\(^1\) Bulk gene expression data from Cell Model Passports (CMP). Corresponding labels are from CREAMMIST, which integrates drug-response curves from GDSC, CCLE, CTRP, and PRISM.}

\newpage
\section{Benchmark details}

\paragraph{Splitting and cross-validation.}
Regression splits are random, but classification splits are stratified by the target. In cohorts with replicates or several samples per patient, splits are additionally grouped by patient, so that no patient contributes to both the support and the query set. In Table \ref{tab:main_results}, classification and regression tasks are evaluated with 5-fold cross-validation and results are averaged across folds, except $R^2$, which is computed from pooled out-of-fold predictions because fold-wise $R^2$ is unstable for small folds. For methods with distinct classification and regression checkpoints, we report the mean parameter count across the two in Table \ref{tab:main_results}.

Survival tasks instead use the first five of the 25 fixed splits distributed with SurvBoard~\citep{wissel2025survboard}. In every split, all data-dependent preprocessing, including standardization and the principal-component projection, is fitted on the support set only.

\paragraph{Model inputs.}
Expression profiles are $\log(1+\mathrm{CPM})$ values over the $G=20{,}021$ reference genes. GeneICL takes these profiles directly and does its own standardization and 90\%-variance PCA. Every other tabular model, including the conventional baselines, gets the same two steps applied before the model. The only exception is TabPFN-Wide, which is designed for inputs with thousands of features and receives $\log(1+\mathrm{CPM})$ values as input.

\paragraph{Transcriptomic foundation models.}
Transcriptomic foundation models are used as frozen feature extractors, without fine-tuning. Each model receives our profiles converted to its expected input format, with genes missing from its vocabulary set to zero, and returns one embedding per sample. These embeddings replace the expression profiles in the pipeline above (standardization and PCA). An elastic-net regularized linear probe is then fitted on the support set of each fold.

COMPASS~\citep{shen2026generalizable}, one of the transcriptomic foundation models we benchmark, expects a cancer-type token for every sample, but it has no token for non-cancer or unknown samples. Because many of our cohorts are not cancer cohorts, every sample receives a neutral token equal to the mean of the 33 learned cancer-type embeddings. We use the self-supervised pretrained checkpoint, not the variants fine-tuned on immunotherapy response, which would add external supervision.

\section{Holdout tasks for model development}
\label{app:holdout}
To avoid overfitting design choices to the benchmark, we ran preliminary experiments on a separate set of datasets, which also served as the validation set monitored during pretraining. The set contains five tasks from five GEO series that are not part of the benchmark and, unlike the benchmark, consists mostly of biological rather than clinically relevant endpoints, such as brain region in post-mortem brain (six classes, GSE102556), age at death from post-mortem brain (GSE80655), age at blood draw (GSE124284), major depressive disorder versus healthy controls (GSE251778), and histological remission in inflammatory bowel disease biopsies (GSE193677, current status, not a prospective label). 

Although these tasks differ from those of our benchmark, conclusions drawn on the holdout tasks transferred to it. Changes that did not affect performance on the holdout tasks (Table~\ref{tab:ablations_holdout}) did not affect performance on the full benchmark (Table~\ref{tab:ablations}) and alternative pretraining distributions lowered regression performance on both sets.

The holdout tasks also show that GeneICL does not benefit from additional input components. Giving the pretrained model all principal components with non-zero variance at inference, rather than only those explaining 90\% of the support-set variance, left classification mostly unchanged (AUROC 0.847 vs.\ 0.846) but lowered regression performance (Spearman 0.691 vs.\ 0.753, pooled $R^2$ 0.506 vs.\ 0.629, see Table~\ref{tab:ablations_holdout}). This supports the 90\% threshold used throughout.

\begin{table}[t]
    \centering
    \caption{
        \textbf{Experiments evaluated on our holdout datasets (three classification and two regression)}. In the last experiment, GeneICL was given all principal components at inference instead of those explaining 90\% of the support-set variance. \emph{Attn.} denotes the asymptotic cost of the feature-axis attention. Parentheses report the standard deviation over three seeds in units of the last decimal place. Best and second-best results are bold and underlined, respectively.
    }
    \vspace{0.5em}
    \scriptsize
    \setlength{\tabcolsep}{3pt}
    \renewcommand{\arraystretch}{1.05}

\begin{tabular}{lcccccccc}
        \toprule
        \textbf{Variant} & \textbf{Params} & \textbf{Attn.}
        & \multicolumn{3}{c}{\textbf{Classification}}
        & \multicolumn{3}{c}{\textbf{Regression}} \\
        \cmidrule(lr){4-6} \cmidrule(lr){7-9}
        & & & \textbf{AUROC} & \textbf{AUPRC} & \textbf{Bal.\ Acc.}
        & \textbf{Pearson} & \textbf{Spearman} & $\mathbf{R^2}$ \\
        \midrule

        \rowcolor{geneiclbluelight}
        GeneICL & 4.24M & $\mathcal{O}(n)$
        & \underline{0.846}(2) & \textbf{0.812}(1) & \textbf{0.787}(10)
        & \textbf{0.806}(7) & \underline{0.753}(8) & \textbf{0.629}(4) \\
        \midrule

        Untied (8 blocks) & 19.25M & $\mathcal{O}(n)$
        & 0.845(1) & \textbf{0.812}(3) & \underline{0.784}(6)
        & \underline{0.805}(5) & \textbf{0.758}(4) & \underline{0.622}(8) \\

        Full attention & 4.05M & $\mathcal{O}(n^2)$
        & 0.843(1) & 0.808(2) & 0.778(11)
        & 0.784(13) & 0.737(20) & 0.581(28) \\

        TabICLv2 prior pretraining & 4.24M & $\mathcal{O}(n)$
        & 0.836(4) & 0.801(5) & 0.769(18)
        & 0.744(11) & 0.688(5) & 0.518(20) \\

        Real expression data pretraining & 4.24M & $\mathcal{O}(n)$
        & 0.843(3) & 0.800(14) & 0.731(59)
        & 0.788(9) & 0.736(8) & 0.457(35) \\

        \midrule

        GeneICL, all PCs at inference & 4.24M & $\mathcal{O}(n)$
        & \textbf{0.847}(2) & \underline{0.811}(5) & 0.772(18)
        & 0.739(12) & 0.691(9) & 0.506(28) \\
        \bottomrule
    \end{tabular}

    \label{tab:ablations_holdout}
\end{table}

\newpage
\section{Hyperparameters}
\begin{table}[h]
    \centering
    \caption{
        \textbf{Hyperparameter selection for GeneICL}. Bolded values indicate identical configurations to TabICLv2.}
    \vspace{0.5em}
    \small
    \setlength{\tabcolsep}{8pt}
    \renewcommand{\arraystretch}{1.1}

    \begin{tabular}{ll}
        \toprule
        \textbf{Parameter} & \textbf{Value} \\
        \midrule
        \midrule
        \multicolumn{2}{l}{\textit{Architecture}} \\
        \midrule
        Embedding dim $D$                                       & \textbf{128} \\
        Attention heads                                        & \textbf{8} \\
        Encoder depth                                          & \textbf{3} \\
        CLS columns                                            & \textbf{4} \\
        Inducing rows (col.\ attn. bottleneck)           & 4 \\
        Thinking rows (scratchpad tokens)$^{1}$                & 8 \\
        Recurrence depth (ICL loops)$^{2}$                     & 8 \\
        Norm                                                   & RMSNorm \\
        Activation function                                    & \textbf{GELU} \\
        Dropout                                                & \textbf{0} \\
        Max classes                                            & \textbf{10} \\
        \midrule
        \multicolumn{2}{l}{\textit{Training}} \\
        \midrule
        Optimizer steps                                        & 20{,}000 \\
        Warmup                                                 & \textbf{5{,}000 steps} \\
        Tasks per step                                         & 6 \\
        Segments$^{3}$                                         & 2 $\times$ 4 steps \\
        Optimizer (FFN matrices)                               & \textbf{Muon}, lr $6\times10^{-4}$, momentum 0.95 \\
        Optimizer (rest)                                       & AdamW ScheduleFree, lr $3\times10^{-4}$ \\
        Weight decay                                           & 0.01 \\
        Gradient clip                                          & \textbf{1.0} \\               
        Seeds                                                  & 0, 1, 2 \\
        \bottomrule
    \end{tabular}
    \label{tab:geneicl_hyperparameters}

    \vspace{0.6em}
    \begin{minipage}{\linewidth}
        \footnotesize
        $^{1}$ Learned reasoning workspace prepended to the ICL context sequence, adapted from TabPFN-2.5.\\
        $^{2}$ Product of $\texttt{segment\_iters}\times\texttt{halt\_max\_segments}$; fixed recurrence unrolling used during inference.\\
        $^{3}$ Hidden state representation $H$ is detached across segment boundaries, with evidence features re-encoded using current parameters at each segment.\\

    \end{minipage}
\end{table}

\section{Computational resource requirements}
\label{sec:cpu-runtime}

\begin{figure}[]
    \centering
    \includegraphics[width=0.6\linewidth]{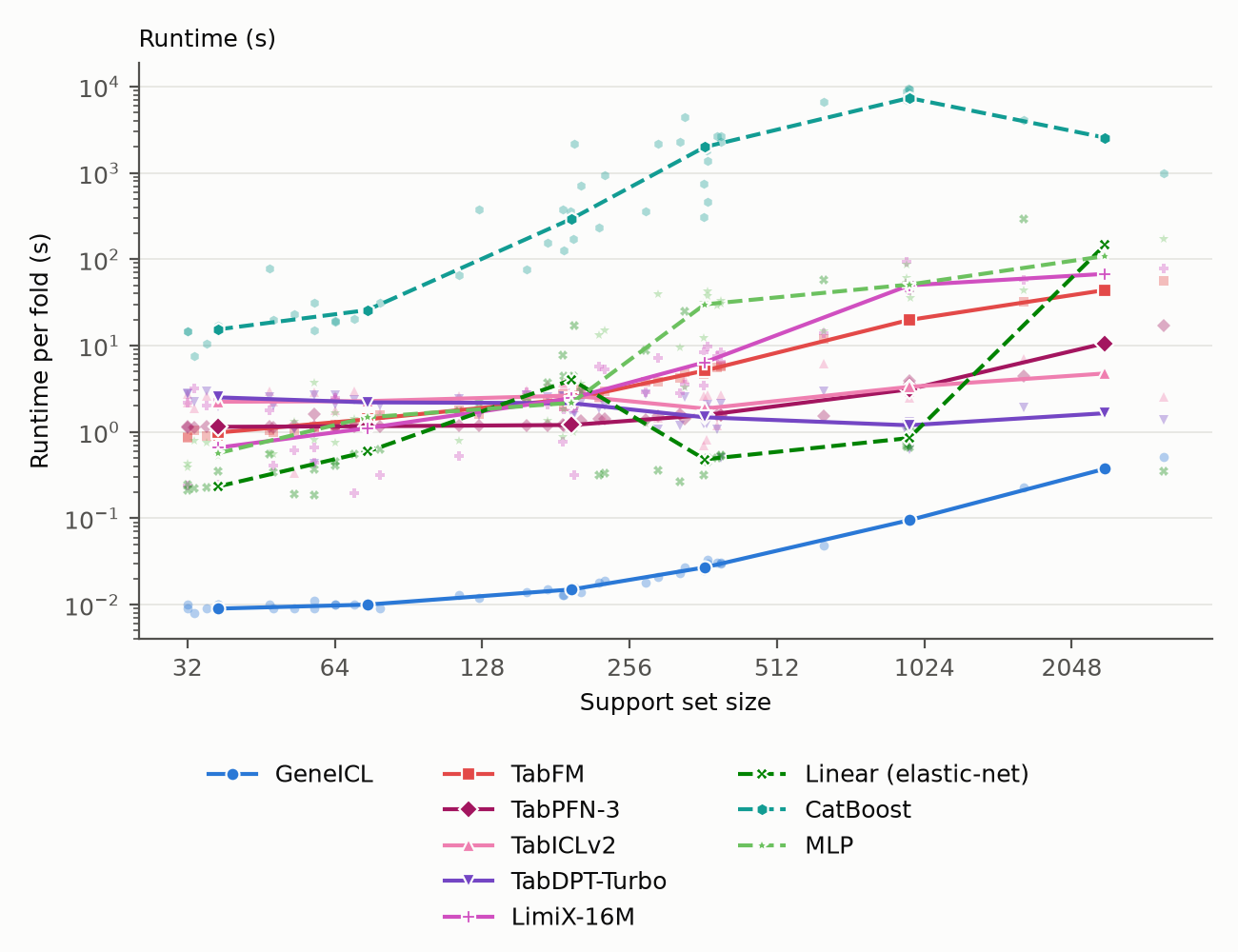}
    \caption{\textbf{Runtime vs. support set size.} Wall-clock runtime per cross-validation fold (log scale) against the number of support (training) samples ($\log_2$ scale), for GeneICL (blue), the other tabular foundation models (red to purple) and the tuned baselines (green, dashed). Each faint marker is one of the 52 benchmark tasks. Lines connect the medians over tasks within support-size bins (edges 32, 64, 128, 256, 512, 1024, 4096). Classification and regression tasks are pooled. Runtime covers fit and predict: for the in-context models that is the forward pass. All models ran on an NVIDIA GeForce RTX 4090 GPU, except CatBoost and the linear model, which ran on AMD EPYC 7763 CPUs.}
    \label{fig:placeholder}
\end{figure}

We recorded the wall-clock runtime per cross-validation fold of all tabular foundation models and tuned baselines during the benchmark (Figure~\ref{fig:placeholder}). For the tuned baselines, this includes the full hyperparameter search (50 candidates, each evaluated with inner 5-fold cross-validation); for the in-context models, it is a single forward pass. Across all support-set sizes, GeneICL is consistently the fastest model, requiring about 0.01--0.4\,s per fold, one to two orders of magnitude less than the other tabular foundation models. Note that CatBoost and the elastic net ran on CPUs, whereas all other models ran on NVIDIA GeForce RTX 4090 GPUs.

In addition, we tested GeneICL without accelerators and ran the classification and regression benchmark (32 datasets) on a laptop CPU (Ryzen~7 7730U, 8~cores, 16~threads) with 16\,GB of RAM. We used the same 5-fold cross-validation protocol as in the main experiments and applied no test-time ensembling. Each fold is one in-context forward pass on the raw gene-expression profiles. Averaged over folds and then over tasks, one fold took \textbf{7.5 seconds} of wall-clock time and the peak memory use was \textbf{4.9 GB}. For comparison, training and inference of an elastic-net baseline took 10.9 seconds per fold on the same laptop, where the average is dominated by the logistic-regression solver on large classification cohorts. GeneICL can therefore be applied to typical bulk transcriptomic cohorts on a standard laptop.

\newpage
\section{Pretraining}

\subsection{Variational autoencoder}
\label{app:vae}

\paragraph{Training data.}
The VAE used for data generation during GeneICL pretraining is trained on human bulk RNA-seq profiles from ARCHS4~\citep{lachmann2018massive} (\texttt{human\_gene\_v2.latest}, upstream build of 5 July 2026). After excluding every GEO series with curated labels and every sample that appears in a benchmark task, we keep samples that are non-single-cell (ARCHS4 single-cell probability below 0.5), have a library size between $5\times10^{5}$ and $10^{8}$ counts, leaving 547{,}829 profiles for VAE training. Counts are converted to $\log(1+\mathrm{CPM})$ using the full-library sum and mapped onto a reference axis of $G=20{,}021$ protein-coding genes (HGNC symbols intersected with GENCODE v44), summing duplicate source symbols and zero-filling reference genes without a match.

\paragraph{Architecture.}
We use a single-latent Gaussian VAE with one hidden layer of width 5{,}000 in both the encoder and the decoder, GELU activations, and a 256-dimensional latent space (see Table~\ref{tab:vae_hyperparameters}). The encoder outputs the mean and log-variance of a diagonal Gaussian posterior. The decoder output is linear and the likelihood is Gaussian with unit variance.

\paragraph{Training.}
Because GeneICL consumes whole datasets rather than individual profiles, the VAE is trained on locally coherent groups of samples. Each step draws $B=8$ anchor profiles uniformly at random and forms a group from each anchor and its $n-1$ nearest neighbors, with $n\sim\mathrm{log} \mathcal{U}[64,512]$ and neighbors found by exact Euclidean search in a 500-component PCA space fitted on a random subsample of 100{,}000 profiles and applied to the whole corpus. Within each group, every gene is standardized across samples. Every row $\mathbf{x}_{bi}\in\mathbb{R}^{G}$ is then encoded independently to a posterior $q_{\psi}(\mathbf{z}\mid\mathbf{x}_{bi})=\mathcal{N}(\boldsymbol{\mu}_{bi},\mathrm{diag}\,\boldsymbol{\sigma}^{2}_{bi})$ and decoded to $\hat{\mathbf{x}}_{bi}=D_{\phi}(\mathbf{z}_{bi})$ with $\mathbf{z}_{bi}\sim q_{\psi}$. In parallel, $m=\min(n,256)$ rows per group are decoded from the prior, $\tilde{\mathbf{x}}_{bj}=D_{\phi}(\tilde{\mathbf{z}}_{bj})$ with $\tilde{\mathbf{z}}_{bj}\sim\mathcal{N}(\mathbf{0},\mathbf{I})$. Writing $\mathbf{X}$ and $\tilde{\mathbf{X}}$ for the real and the prior-decoded batch, the loss at step $t$ is
\begin{equation}\label{eq:vae_loss}
\mathcal{L}_{\mathrm{VAE}}
=
\mathcal{L}_{\mathrm{rec}}
+
\beta_{t}\,\mathcal{L}_{\mathrm{KL}}
+
\lambda_{\mathrm{cov}}\,\mathcal{L}_{\mathrm{cov}}
+
\lambda_{\mathrm{mm}}\,\mathcal{L}_{\mathrm{mm}},
\end{equation}
where $\beta_{t}=\min(1,\,t/1{,}500)$ implements the KL warm-up and $\lambda_{\mathrm{cov}}=20$, $\lambda_{\mathrm{mm}}=10$. The first two terms are the usual ELBO under a unit-variance Gaussian likelihood,
\begin{equation}\label{eq:vae_elbo}
\mathcal{L}_{\mathrm{rec}}
=
\frac{1}{Bn}\sum_{b,i}\tfrac{1}{2}\bigl\|\hat{\mathbf{x}}_{bi}-\mathbf{x}_{bi}\bigr\|_{2}^{2},
\qquad
\mathcal{L}_{\mathrm{KL}}
=
\frac{1}{Bn}\sum_{b,i}D_{\mathrm{KL}}\bigl(q_{\psi}(\mathbf{z}\mid\mathbf{x}_{bi})\,\big\|\,\mathcal{N}(\mathbf{0},\mathbf{I})\bigr).
\end{equation}
The two auxiliary terms see only prior samples and constrain dataset-level structure that the ELBO leaves free,
\begin{equation}\label{eq:vae_aux}
\mathcal{L}_{\mathrm{cov}}
=
1-\mathrm{corr}\Bigl(\mathrm{offdiag}\,\hat{\boldsymbol{\Sigma}}_{\mathcal{S}}(\tilde{\mathbf{X}}),\;\mathrm{offdiag}\,\hat{\boldsymbol{\Sigma}}_{\mathcal{S}}(\mathbf{X})\Bigr),
\qquad
\mathcal{L}_{\mathrm{mm}}
=
\Bigl(\log \bar{v}(\tilde{\mathbf{X}})-\log \bar{v}(\mathbf{X})\Bigr)^{2}.
\end{equation}
Here $\mathcal{S}$ is a fresh random subset of 128 genes at every step, $\hat{\boldsymbol{\Sigma}}_{\mathcal{S}}$ is the gene--gene covariance over $\mathcal{S}$ obtained by centering each group on its own mean and pooling the groups, $\mathrm{corr}$ is the Pearson correlation over the upper-triangular entries, and $\bar{v}(\cdot)$ is the within-group variance of a gene, averaged over genes and groups. We train for 10{,}000 steps with AdamW on a single RTX 4090 (about 70 minutes) and use the final checkpoint. At the end of training, generated and real gene--gene covariances correlate at $r=0.87$, measured on 32 evaluation neighborhoods over 256 randomly chosen genes, and generated samples retain 53\% of the real within-group variance.

\paragraph{Sampling.}
During GeneICL pretraining, every row of a synthetic dataset is generated independently by drawing $\mathbf{z}\sim\mathcal{N}(\mathbf{0},\mathbf{I})$ and taking the decoder mean $D_\phi(\mathbf{z})$, without observation noise or temperature scaling.

\begin{table}[h]
\centering
\caption{\textbf{VAE configuration and hyperparameters.}}
\label{tab:vae_hyperparameters}
\begin{tabular}{ll}
\toprule
\textbf{Setting} & \textbf{Value} \\
\midrule
Input dimension & 20{,}021 genes \\
Hidden width (encoder / decoder) & 5{,}000 / 5{,}000 \\
Latent dimension & 256 \\
Activation & GELU \\
Parameters & 204M \\
KL weight & 1, linear warm-up over 1{,}500 steps \\
Auxiliary loss weights (covariance / moment matching) & 20 / 10 \\
Groups per step $\times$ group size & $8 \times n$, $n\sim\mathrm{logU}[64,512]$ \\
Optimizer & AdamW, lr $2\times10^{-4}$, weight decay $10^{-4}$ \\
Gradient clipping & 1.0 \\
Training steps & 10{,}000 \\
\bottomrule
\end{tabular}
\end{table}

\subsection{Pretraining prior}
\label{app:pretraining_prior}

Table~\ref{tab:prior} lists the sampling distributions of the semi-synthetic prior. We write log $\mathcal{U}[a,b]$ for a log-uniform distribution and log $\mathcal{U}\{a,\ldots,b\}$ for its integer counterpart $\lfloor\exp(\mathcal{U}[\log a, \log(b+1)])\rfloor$, which places probability proportional to $\log\frac{k+1}{k}$ on each integer $k$.

\paragraph{Task and label space.}
Every training batch contains six synthetic datasets that share a sampled number of rows $n$ and support fraction. Labels are constructed in a per-dataset principal-component representation of the generated matrix $\mathbf{X}$: we retain the leading components explaining 90\% of the variance, whiten each score to unit variance, and clip it to $[-10,10]$. Each score function then reads a subset $\mathcal{J}$ of $k$ of these components, drawn without replacement with probability proportional to their explained-variance ratio, so that task signal concentrates on prominent axes of transcriptomic variation while occasionally involving minor ones.

\paragraph{Score functions.}
Each latent score $s_{ik}$ is the average of an ensemble of one to three independently sampled MLPs, each standardized over samples before averaging. Every weight row is generated by applying a softmax to $-\alpha\log j + \beta\,\varepsilon_j$ ($j=1,\ldots,d_{\mathrm{in}}$, $\varepsilon_j\sim\mathcal{N}(0,1)$), randomly permuting and sign-flipping the entries, and rescaling to norm $\sqrt{d_{\mathrm{in}}}$. Small $\alpha$ and $\beta$ yield dense, evenly spread weights, whereas large values concentrate the weight on a few inputs. Hidden pre-activations are standardized, scaled by a random pre-activation scaling, passed through a randomly chosen nonlinearity, and standardized again. In small-scale experiments, adding non-MLP score functions such as random decision-tree ensembles did not improve downstream performance. Richer function classes nonetheless remain a promising direction for future work.

\paragraph{Labels.}
Each latent score is mixed with independent Gaussian noise to reach a sampled signal fraction $t$ (the fraction of target variance explained by the score). Binary labels threshold the first score at a random quantile, multiclass labels take the argmax over $K$ scores, and the regression target is the first score. With probability 0.2, the regression target is additionally floored and/or capped at random quantiles, mimicking detection limits and saturated clinical measurements. Regression targets are standardized with support-set statistics.

\begin{table}[h]
\centering
\caption{\textbf{Sampling distributions of the pretraining prior.} Quantities are resampled for every synthetic dataset unless marked as shared across the batch.}
\vspace{4pt}
\label{tab:prior}
\small
\begin{tabular}{p{0.38\linewidth}p{0.55\linewidth}}
\toprule
\textbf{Quantity} & \textbf{Distribution} \\
\midrule
\multicolumn{2}{l}{\emph{Task and label space}} \\
Support fraction (shared)
& $\mathcal{U}[0.5, 0.9]$ \\
Samples per task                    & log $\mathcal{U}[32,1024]$ \\
Number of selected PCs $k$ & log $\mathcal{U}\{2,\ldots,\min(64, k_{\mathrm{tot}})\}$ \\
\midrule
\multicolumn{2}{l}{\emph{Labels}} \\
Number of classes $K$ & log $\mathcal{U}\{2,\ldots,10\}$ \\
Binary threshold quantile & $\mathcal{U}[0.2, 0.8]$ \\
\midrule
\multicolumn{2}{l}{\emph{Random MLP functions}} \\
Ensemble size & $\mathcal{U}\{1,2,3\}$ \\
Number of layers & 1 / 2 / 3 with probability 0.2 / 0.5 / 0.3 \\
Hidden width & log $\mathcal{U}\{4,\ldots,64\}$ \\
Weight concentration exponent $\alpha$ & log $\mathcal{U}[0.1/\log(1+d_{\mathrm{in}}), 6]$ \\
Weight noise $\beta$ & log $\mathcal{U}[10^{-4}, 10]$ \\
Bias & log $\mathcal{U}[0.01, 1]\cdot\mathcal{N}(0,1)$ \\
Pre-activation scaling & log $\mathcal{U}[0.5, 3]$ \\
\bottomrule
\end{tabular}
\end{table}

\newpage
\section{Survival prediction}
\label{app:survival}

This section details GeneICL-Cox and the three groups of methods it is compared against: four alternative censoring adaptations applied to the same frozen GeneICL backbone, three tuned conventional survival models, and SurvPFN.

\paragraph{GeneICL-Cox.}
Let $(t_i,\delta_i)$ denote the observed time and event indicator of support sample $i$. We first transform the support labels into the martingale residuals $r_i$ of Eq.~\ref{eq:martingale}, which requires no model, and then regress them with GeneICL in a single pass. Support scores are predicted out-of-fold over five random folds and query scores from the full support set: the out-of-fold scores are what the Breslow baseline hazard $\hat{H}_0$ is subsequently estimated at, and in-sample scores would be over-separated and bias $\hat{H}_0$ downwards, decalibrating the survival curves $S(t\mid\mathbf{x})=\exp(-\hat{H}_0(t)\,e^{f(\mathbf{x})})$. 

\paragraph{Alternative censoring adaptations.}
The following four methods use the same frozen GeneICL backbone as GeneICL-Cox and differ only in how right-censored outcomes are turned into targets. The first three are the adaptations evaluated by \citet{vieyra2026staying}, who introduced the Buckley--James and jackknife variants and adopted the discrete-time variant from \citet{kim2026tabular}. Two-step imputation is a simple imputation baseline.
\begin{itemize}[leftmargin=*]
    \item \textbf{Buckley--James estimator}: This method iteratively imputes censored log-times. In each of 10 iterations, GeneICL regresses log-time using all events and a random half of the censored samples as context. Each censored log-time is then re-imputed by sampling from a log-normal residual distribution truncated below at the censoring time, and the scale $\sigma$ is re-estimated by censored maximum likelihood. Imputations and $\sigma$ are initialized from the jackknife variant below.
    \item \textbf{Jackknife pseudo-observations}: Censored targets are replaced by leave-one-out Kaplan--Meier pseudo-observations of survival at the median event time, which are converted into survival-time pseudo-targets. GeneICL then regresses these targets in a single pass.
    \item \textbf{Discrete-time hazards}: The time axis is partitioned into 10 intervals at percentiles of the training event times. For each interval, GeneICL's classification head predicts the conditional event probability $h_k$ among samples still at risk. The survival function is then $S(t)=\prod_{k:\,\tau_k\le t}(1-h_k)$.
    \item \textbf{Two-step imputation}: GeneICL first predicts event times for the censored samples using only the uncensored samples as context. It then predicts query event times using all samples, with the imputed times, as context. The resulting survival function is a step function at the predicted event time.
\end{itemize}

\paragraph{Baseline methods.}
\begin{itemize}[leftmargin=*]
    \item \textbf{Elastic net}: Cox PH elastic net was implemented using \texttt{glmnet}~\citep{glmnet, glmnet_cox} in \texttt{R}, using a fixed $\alpha$ mixing component of 0.1 (\texttt{alpha=0.1}), five-fold inner cross-validation stratified on the event indicator, and considering 100 regularization hyperparameters (\texttt{nlambda=100}), among which the hyperparameter yielding minimum cross-validation error was chosen (\texttt{lambda.min}). All other parameters were left at their respective defaults. Survival functions were extracted using \texttt{pec} at all unique test timepoints.
    \item \textbf{Random survival forests}: Random survival forests~\citep{rsfs} were implemented using \texttt{scikit-survival}~\citep{polsterl2020scikit} in \texttt{Python}, using inner five-fold cross-validation stratified on the event indicator to choose the set of hyperparameters maximizing mean Harrell's concordance based on the ensemble risk score prediction for each set of hyperparameters. Mirroring~\citet{kim2026tabular}, we tuned \texttt{n\_estimators} $\in \{50, 100, 200\}$, and \texttt{min\_samples\_split} $\in \{2, 5, 10, 20\}$. All other parameters were left at their respective defaults. Survival functions were extracted using the \texttt{predict\_survival\_function} at all unique test timepoints, clipped to the training time domain.
    \item \textbf{XGBoost}: A Cox PH gradient boosting model was implemented using \texttt{xgboost}~\citep{xgb} in \texttt{Python} (\texttt{survival:cox}). Hyperparameters were selected via inner five-fold cross-validation stratified on the event indicator, using \texttt{GridSearchCV}. Following~\citet{kim2026tabular}, we tuned \texttt{n\_estimators} $\in \{100, 200\}$, \texttt{learning\_rate} $\in \{0.1, 0.3\}$, and \texttt{max\_depth} $\in \{3, 6\}$. For model selection, the cross-validation score was the mean negative Cox partial log-likelihood returned directly by XGBoost (\texttt{cox-nloglik}), obtained by calling the fitted booster evaluation API (\texttt{Booster.eval\_set}) on each validation fold. All other parameters were left at their respective defaults. Survival functions were obtained by fitting a Breslow baseline hazard from \texttt{scikit-survival}~\citep{polsterl2020scikit} to the predicted linear predictors and then evaluating the resulting survival step functions at all unique test timepoints, with evaluation times clipped to the training-time domain.
\end{itemize}

\paragraph{SurvPFN.} In addition to tuned baselines, we compare GeneICL to SurvPFN~\citep{bohm2026survpfn}, a tabular foundation model pretrained on synthetic right-censored survival data that predicts a discretized distribution over log survival times. We used the released checkpoint with the full training split as context and the same input as GeneICL, namely the principal components explaining 90\% of the variance, fitted on the training split. Risk scores were the negative predicted mean log survival time, and survival functions were evaluated at all unique test timepoints.

\paragraph{Metrics.}
Discriminative performance is measured using Antolini's time-dependent concordance index~\citep{antolini2005time}, an overall score incorporating both discrimination and calibration via the integrated Brier score (IBS) using 100 points~\citep{graf1999assessment}, and distributional calibration through the D-calibration test statistic using 10 bins~\citep{haider2020effective}. All metrics are calculated using \texttt{SurvivalEVAL}~\citep{qi2023survivaleval} on the first five SurvBoard test splits and averaged per dataset. 

Predicted survival curves were post-processed to account for numerical issues in selected models (primarily enforcing monotonic non-increasing survival over time via carrying forward the survival probability at the previous timepoint where applicable). 

D-calibration is undefined for the \texttt{two-step} variant and is thus not reported. XGBoost sometimes resulted in survival functions on which D-calibration calculation did not succeed, in which case these splits were reported as NA and subsequently imputed using the mean D-calibration across splits for the same model and the same dataset.

\newpage
\section{Extended ablations}

\subsection{Positional encodings and feature grouping}
\label{app:arch_symmetry}
TabICL and TabICLv2 use rotary positional encodings and circular feature grouping to break the symmetry between features with similar distributions, without which samples that differ only by a permutation of their feature values can collapse to the same representation~\citep{qu2026tabiclv2}. GeneICL omits both because principal components have no order or sign that is shared across datasets, continuous PC scores rarely coincide under a permutation, and the support labels added in the encoder already distinguish features by their relation to the target.

\subsection{Looping}
\label{app:arch_looping}
\begin{figure}[t]
    \centering

    \includegraphics[width=0.45\linewidth]{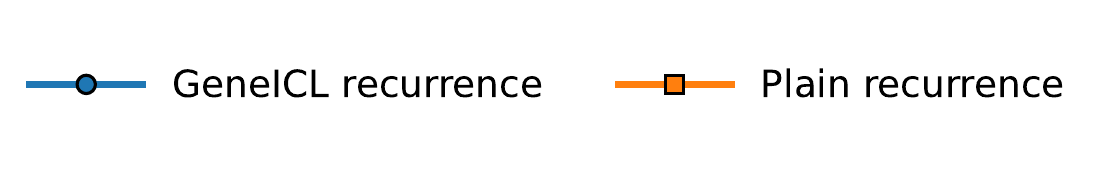}

    \vspace{0.4em}

    \includegraphics[width=0.32\linewidth]{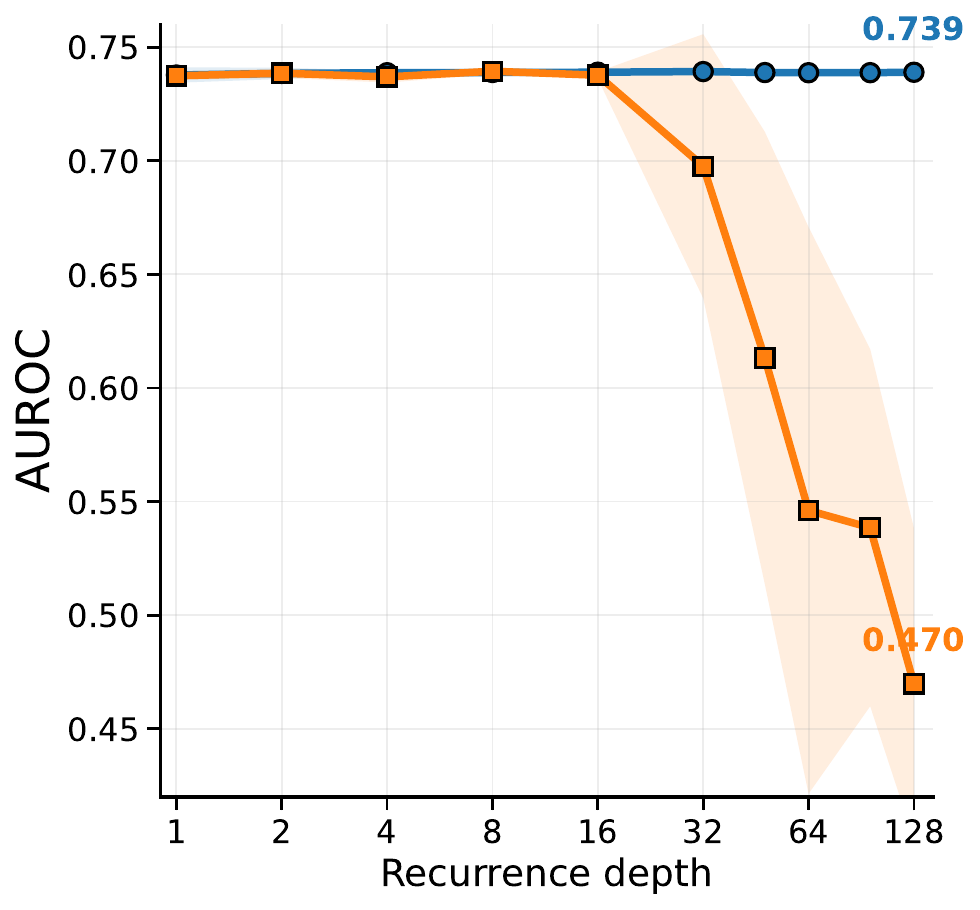}
    \hfill
    \includegraphics[width=0.32\linewidth]{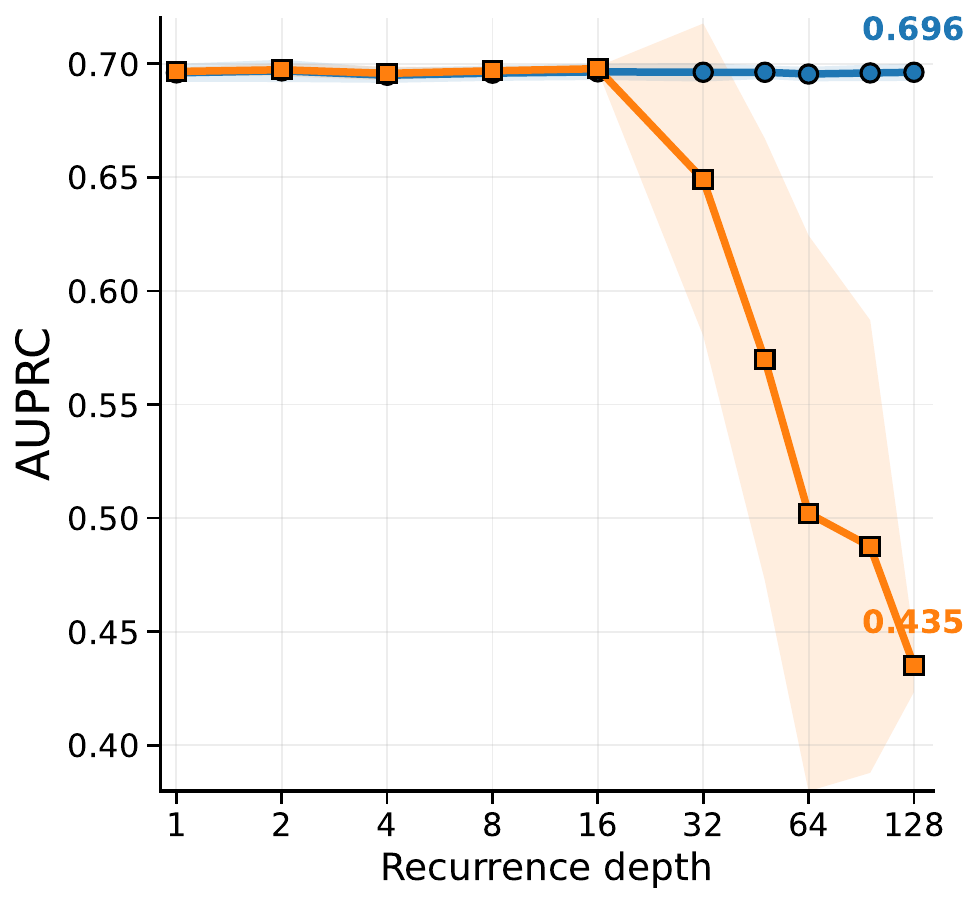}
    \hfill
    \includegraphics[width=0.32\linewidth]{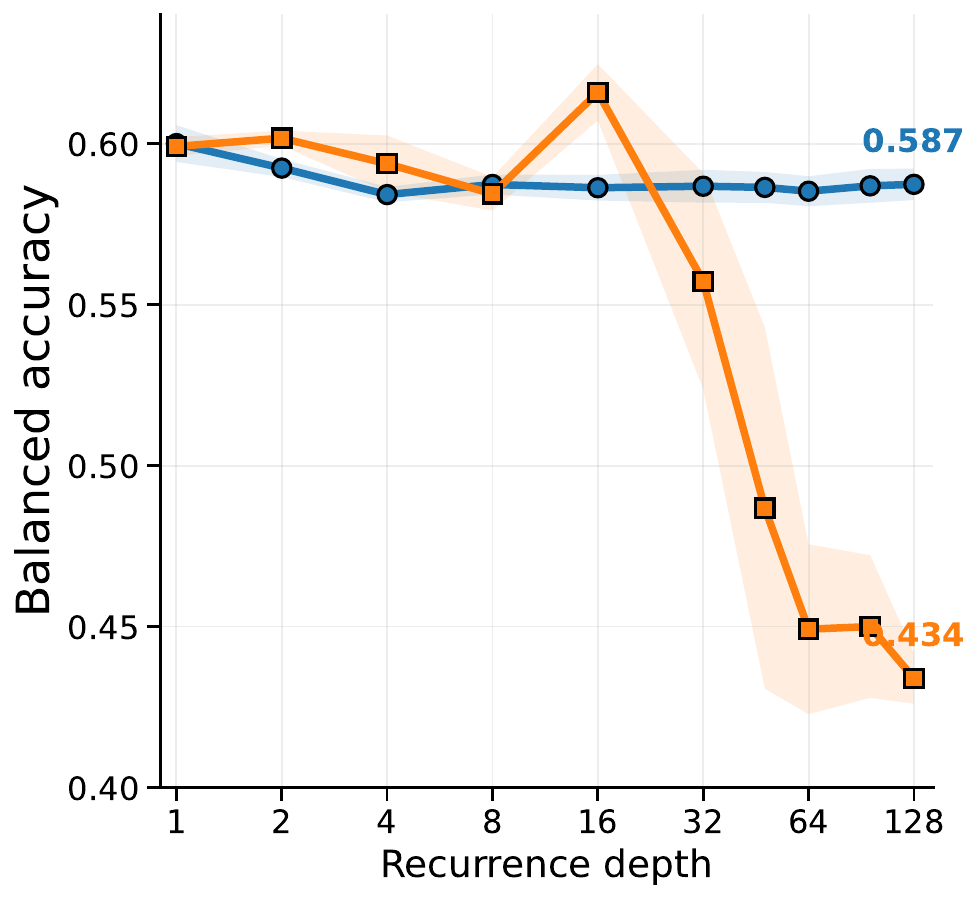}

    \vspace{0.4em}

    \includegraphics[width=0.32\linewidth]{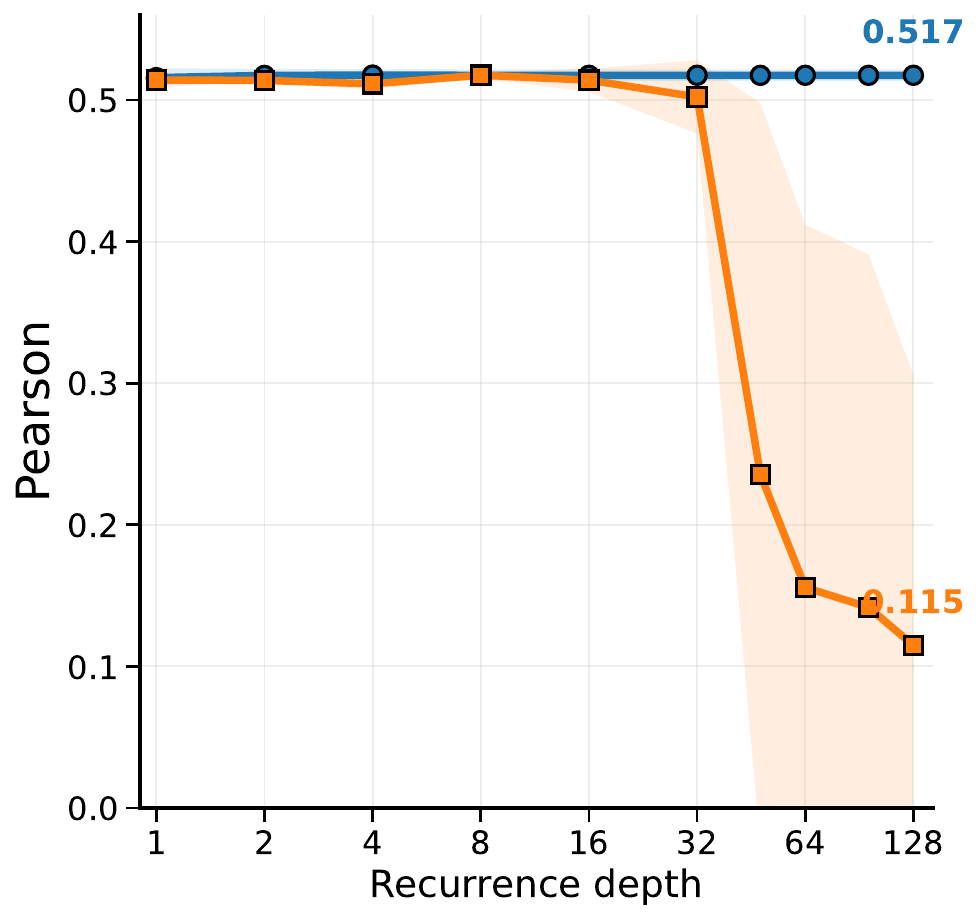}
    \hfill
    \includegraphics[width=0.32\linewidth]{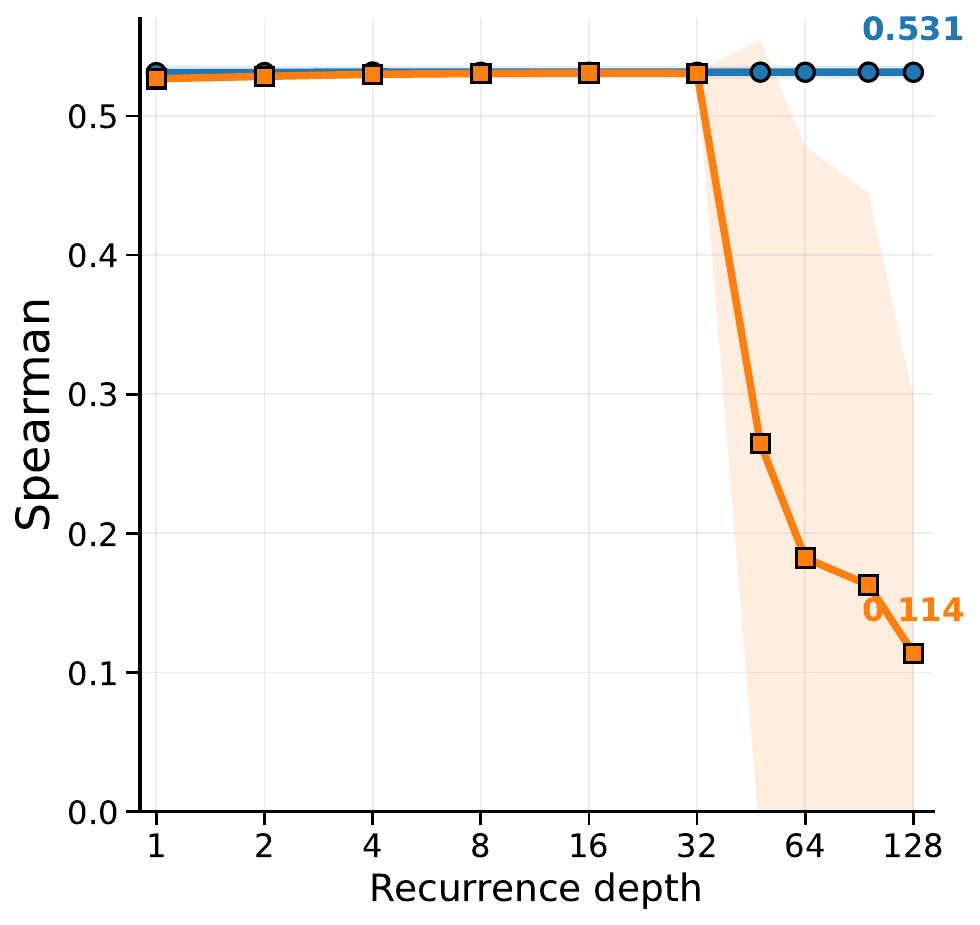}
    \hfill
    \includegraphics[width=0.32\linewidth]{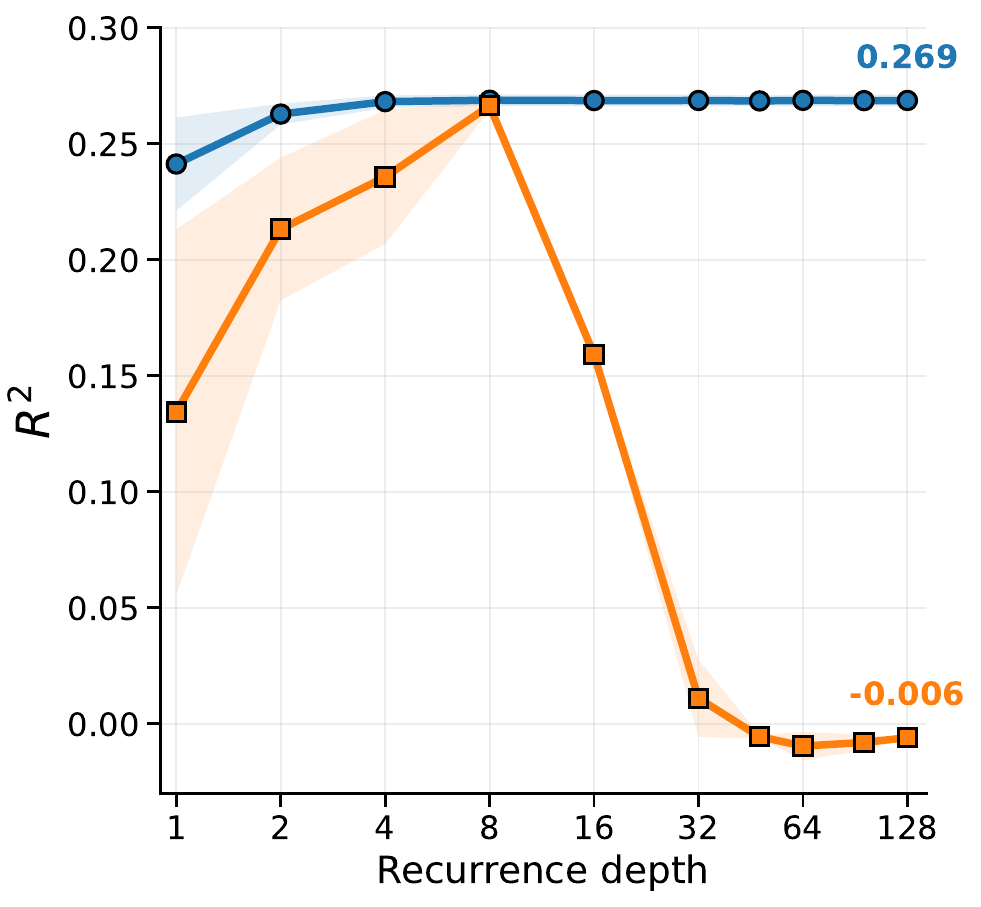}

    \caption{
        \textbf{Depth extrapolation of the recurrent in-context learner.}
        GeneICL and a plain recurrent baseline are trained with $T=8$ recurrence
        steps and evaluated with up to 128 steps at inference time.
        The plain baseline repeatedly applies a post-norm Transformer block, whereas
        GeneICL uses the fixed-evidence, evolving-hidden-state recurrence described in
        Section~\ref{sec:geneicl_architecture}.
        Panels report \textbf{(a)} AUROC, \textbf{(b)} AUPRC,
        \textbf{(c)} balanced accuracy, \textbf{(d)} Pearson correlation,
        \textbf{(e)} Spearman correlation, and \textbf{(f)} $R^2$.
        Lines and markers show the mean over three independently trained seeds and
        shaded regions indicate $\pm1$ standard deviation.
        GeneICL remains stable far beyond the training depth, whereas plain recurrence
        degrades under continued iteration.
    }
    \label{fig:arch_looping}
\end{figure}
GeneICL is trained with $T=8$ recurrence steps. To test whether its recurrent
computation remains stable beyond the depth encountered during training, we
evaluate pretrained checkpoints at
$T \in \{1,2,4,8,16,32,48,64,96,128\}$.
We compare the GeneICL recurrence described in Section~\ref{sec:geneicl_architecture},
which separates fixed task evidence from an evolving hidden state, against a
plain recurrent baseline that repeatedly applies a post-norm Transformer block
without these recurrent design choices. Both models are trained with the same
recurrence depth of $T=8$. Results are averaged over three independently trained
seeds.

Performance generally reaches its maximum, or a near-maximum, within the first
few recurrence steps (Figure~\ref{fig:arch_looping}). Beyond this point,
additional iterations do not systematically improve GeneICL. Crucially,
however, its predictions remain stable as recurrence depth is increased far
beyond the training horizon. Across all six classification and regression
metrics, performance at $T=128$ remains essentially unchanged from that around
the training depth. Plain recurrence behaves differently. While it is
competitive at small depths, performance deteriorates once the block is
repeated substantially beyond the depth seen during training, with particularly
large declines in Pearson correlation, Spearman correlation, and $R^2$.

Thus, weight sharing by itself is not sufficient for stable recurrent depth
extrapolation. GeneICL's recurrent formulation supports substantially deeper
iteration without destabilizing predictive performance, although additional
recurrence does not provide a performance benefit on the present benchmark.
One possible reason is that current biomedical prediction tasks may be limited
more by the information contained in the measurements than by available model
computation. Recent work argues that measurement noise can preferentially
attenuate nonlinear predictive structure, thereby reducing the advantage of
more flexible predictors~\citep{schulz2026measurement}. Whether additional
recurrent computation becomes beneficial as transcriptomic measurements become
more precise is therefore an interesting direction for future work.

\subsection{Inference-time ensemble}
\label{app:arch_ensemble}
\begin{figure}[t]
    \centering
    \includegraphics[width=0.62\linewidth]{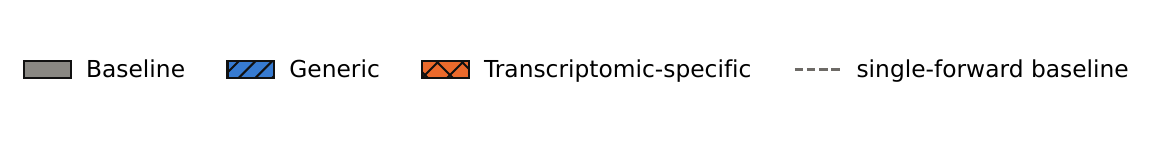}
    \includegraphics[width=\linewidth]{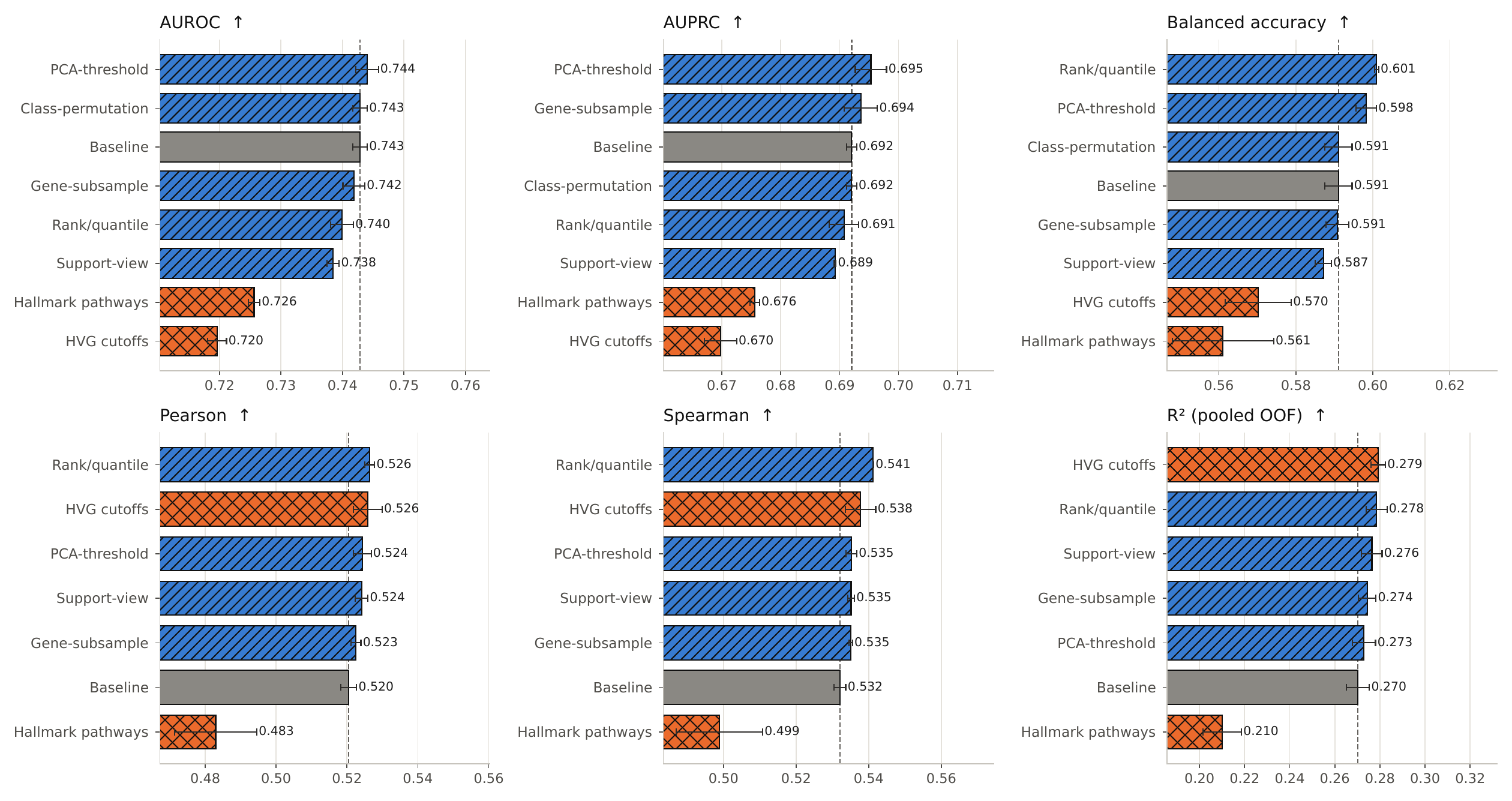}
    \caption{
        \textbf{Inference-time ensembling strategies for GeneICL.}
        Each ensembling method is applied independently to a frozen GeneICL checkpoint and compared against a single-forward-pass baseline.
        We report AUROC, AUPRC, and balanced accuracy for classification, and Pearson correlation, Spearman correlation, and $R^2$ for regression.
        Bars show the mean performance across three independently trained seeds, with error bars indicating one standard deviation.
        The dashed vertical line marks the single-forward baseline.
        Generic ensembling strategies are compared with transcriptomic-specific strategies based on highly variable genes and Hallmark pathways.
        Higher is better for all metrics.
    }
    \label{fig:ensembling_methods}
\end{figure}

We investigate whether GeneICL can benefit from inference-time ensembling without retraining or updating model parameters. Each strategy constructs multiple views of the same prediction problem using a single frozen GeneICL checkpoint and averages the resulting class probabilities for classification or point predictions for regression. We evaluate all strategies using three independently trained checkpoints and compare them against a single-forward baseline using the default PCA threshold of $0.90$. The goal is to test whether robustness can be improved by varying the input representation or support set, and whether transcriptomics-specific structure provides additional benefits over generic ensembling strategies.

We consider five generic ensembling strategies. \emph{PCA-threshold} ensembling evaluates the model using PCA variance-retention thresholds of $\{0.85, 0.90, 0.95, 0.98\}$ and averages the resulting predictions, thereby combining multiple low-dimensional representations of the same expression profiles. \emph{Support-view} ensembling averages predictions over eight class-stratified subsets containing $90\%$ of the available support examples. \emph{Gene-subsample} ensembling similarly averages over eight random subsets containing $50\%$ of the input genes, providing a feature-bagging analogue for transcriptomic data. \emph{Rank/quantile}~\citep{bolstad2003comparison} ensembling combines three different distributional representations of the expression matrix: the original values, a per-sample rank transformation, and quantile normalization to the mean sorted expression profile. Finally, \emph{class-permutation} ensembling, which applies only to classification, averages predictions over eight permutations of the mapping between class labels and class indices after mapping the outputs back to the original label space.

We additionally evaluate two transcriptomics-specific alternatives. \emph{HVG-cutoff} ensembling restricts the input to the top $500$, $1{,}000$, $2{,}000$, or $5{,}000$ highly variable genes~\citep{brennecke2013accounting}, with variability estimated from the support set, and averages predictions across the four resulting views. \emph{Hallmark-pathway} ensembling instead evaluates the model separately on each of the 50 MSigDB Hallmark gene sets~\citep{liberzon2015molecular} and averages the corresponding predictions. These two approaches test whether ensembles based on biologically motivated feature subsets provide an advantage over generic perturbations of the input representation.

As shown in Figure~\ref{fig:ensembling_methods}, the generic ensembling strategies are generally competitive with the single-forward baseline and provide modest improvements for several metrics. PCA-threshold, support-view, gene-subsample, and rank/quantile ensembling all preserve or improve performance on at least part of the benchmark, with the strongest strategy depending on the metric. In contrast, the transcriptomics-specific approaches do not provide consistent gains. HVG-based ensembling substantially reduces classification performance, while Hallmark-pathway ensembling performs worse across all reported metrics. These results indicate that GeneICL benefits more from generic variation in the representation or support set than from restricting inference to predefined transcriptomic feature subsets.

\end{document}